\documentclass[10pt,twocolumn,letterpaper]{article}

\usepackage{cvpr}              % To produce the CAMERA-READY version
\usepackage[dvipsnames]{xcolor}

\definecolor{cvprblue}{rgb}{0.21,0.49,0.74}
\usepackage[pagebackref,breaklinks,colorlinks,citecolor=cvprblue]{hyperref}
\usepackage{colortbl}
\usepackage{multicol}
\usepackage{multirow}

\usepackage{pifont}% http://ctan.org/pkg/pifont
\newcommand{\xmark}{\textcolor{red}{\ding{55}}}%
\newcommand{\cmark}{\textcolor{green}{\ding{51}}}%

\usepackage{tikz}
\def\paperID{719} % *** Enter the Paper ID here
\def\confName{CVPR}
\def\confYear{2024}

\title{CapHuman: Capture Your Moments in Parallel Universes}

\author{Chao Liang$^{1}$ \qquad
    Fan Ma$^{1}$ \qquad
    Linchao Zhu$^{1}$$^\dag$ \quad
    Yingying Deng$^{2}$ \quad
    Yi Yang$^{1}$ \\
    $^{1}$ReLER, CCAI, Zhejiang University \quad
    $^{2}$Huawei Technologies Ltd. \\
    {\small$^\dag$ Corresponding author} \\
    {\tt\small {\{cs.chaoliang, zhulinchao, yangyics\}}@zju.edu.cn, flower.fan@foxmail.com, dyy15@outlook.com} \\
    \url{https://caphuman.github.io}
}
\begin{document}
\twocolumn[{%
\renewcommand\twocolumn[1][]{#1}%
\maketitle
\begin{tikzpicture}[remember picture,overlay,shift={(current page.north west)}]
\node[anchor=north west,xshift=2.0cm,yshift=-2.9cm]{\scalebox{-1}[1]{\includegraphics[width=2.0cm]{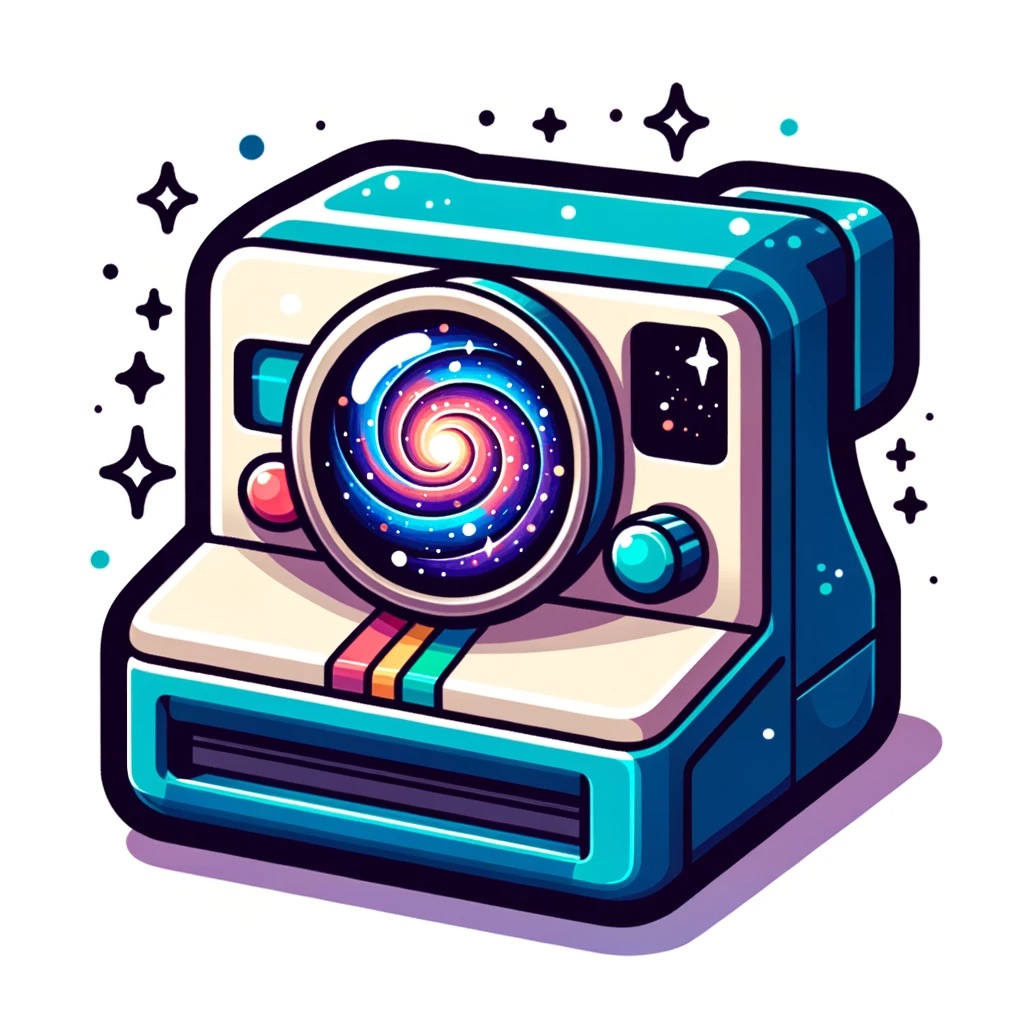}}};
\end{tikzpicture}
\vspace{-3.0em}
% \vspace{-3.8em}
\begin{center}
    \centering
    \captionsetup{type=figure}
    \includegraphics[width=0.98\textwidth]{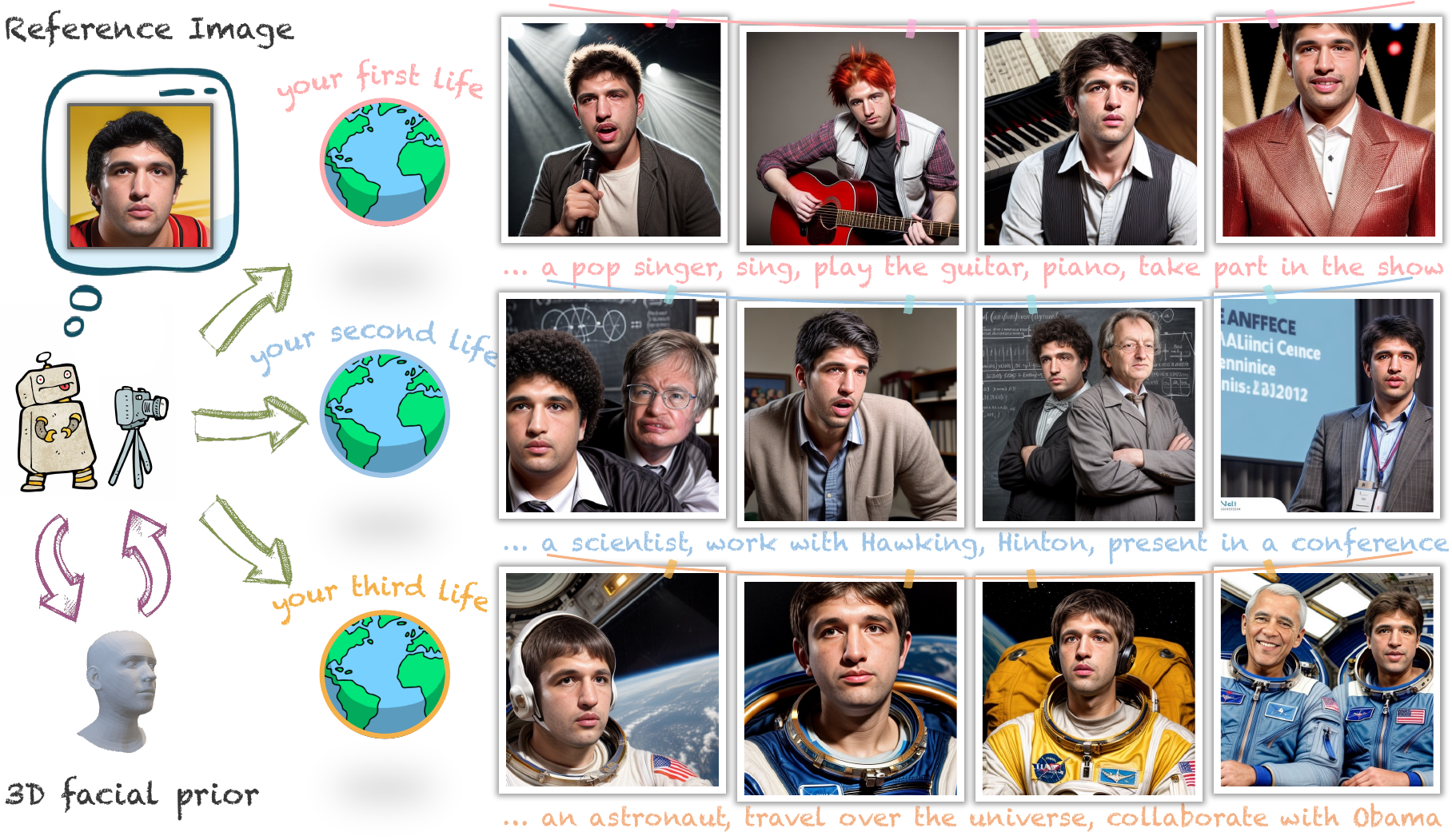}
    % \vspace{-2.3em}
    \vspace{-1.0em}
    \captionof{figure}{Given only one reference facial photograph, our CapHuman can generate photo-realistic specific individual portraits with content-rich representations and diverse head positions, poses, facial expressions, and illuminations in different contexts.
    }
    \label{fig:teaser}
\end{center}%
}]

% \blfootnote{\small$^\dag$ Corresponding author.}

\begin{abstract}
% \vspace{-1.2em}
We concentrate on a novel human-centric image synthesis task, that is, given only one reference facial photograph, it is expected to generate specific individual images with diverse head positions, poses, facial expressions, and illuminations in different contexts. To accomplish this goal, we argue that our generative model should be capable of the following favorable characteristics: (1) a strong visual and semantic understanding of our world and human society for basic object and human image generation. (2) generalizable identity preservation ability. (3) flexible and fine-grained head control. Recently, large pre-trained text-to-image diffusion models have shown remarkable results, serving as a powerful generative foundation. As a basis, we aim to unleash the above two capabilities of the pre-trained model. In this work, we present a new framework named CapHuman. We embrace the ``encode then learn to align" paradigm, which enables generalizable identity preservation for new individuals without cumbersome tuning at inference. CapHuman encodes identity features and then learns to align them into the latent space. Moreover, we introduce the 3D facial prior to equip our model with control over the human head in a flexible and 3D-consistent manner. Extensive qualitative and quantitative analyses demonstrate our CapHuman can produce well-identity-preserved, photo-realistic, and high-fidelity portraits with content-rich representations and various head renditions, superior to established baselines. Code and checkpoint will be released at \url{https://github.com/VamosC/CapHuman}.
\end{abstract}    
\section{Introduction}
\label{sec:intro}
\begin{quote}
\it \small
John Oliver:``$\cdots$ Does that mean there is a universe out there where I am smarter than you?" \\
Stephen Hawking:``Yes. And also a universe where you're funny." \\
\mbox{}\hfill -- Last Week Tonight 
\vspace{-10pt}
\end{quote}

There are infinite possibilities in parallel universes. The parallel universe, \ie multiverse, is a many-worlds interpretation of quantum mechanics. When mapping into the realism framework, it means there might be thousands of different versions of our lives out here, living simultaneously. Our human beings are naturally imaginative. We are strongly eager for our second life to play different roles that have never been explored yet. Have you ever dreamed that you are a pop singer in the spotlight? Have you ever dreamed that you become a scientist, working with Stephen Hawking and Geoffrey Hinton? Or, have you ever dreamed that you act as an astronaut and have a chance to travel around the vast universe fearlessly? It will be quite satisfactory to capture our different moments in parallel universes if possible. To make our dreams come true, we raise an open question: can we resort to the help of the current machine intelligence and is it ready?

Thanks to the rapid development of advanced image synthesis technology in generative models~\cite{ramesh2021dalle, ramesh2022dalle2, nichol2021glide, rombach2022SD, saharia2022imagen, zhou2023pyramid}, the recent large text-to-image diffusion models bring the dawn of possibilities. They show promising results in generating photo-realistic, diverse, and high-quality images. To achieve our goal, we first analyze and decompose the fundamental functionalities of our model. In our scenario~(see Figure~\ref{fig:teaser}), an ideal generative model should have the following favorable properties: (1) \textit{a strong visual and semantic understanding of our world and human society}, which can provide the basic capabilities of object and human image generation. (2) \textit{generalizable identity preservation ability}. Identity information is often described as a kind of visual content. It is represented as even only one reference photograph in some extreme situations, in order to meet the user's preference. This requires our generative model to learn to extract key identity features, well-generalizable to new individuals. (3) \textit{flexible to put the head everywhere with any poses and expressions in fine-grained control}. Human-centric image generation demands our model to support the geometric control of facial details. Then, we dive deep into the existing methods and investigate their availability. Poorly, all of them cannot meet all the aforementioned requirements. On the one hand, a number of works~\cite{gal2023TI, ruiz2023dreambooth, hu2021lora} attempt to personalize the pre-trained text-to-image model by fine-tuning at test-time, suffering from the overfitting problem in the one-shot setting. They are insufficient to supply the head control as well. On the other hand, some works~\cite{zhang2023controlnet, mou2023t2iadapter, ding2023diffusionrig} focus on the head control. However, these approaches cannot preserve the individual identity or are trained from scratch without a good vision foundation and lack of text control, so as to constrain their generative ability. 

In this work, we propose a novel framework CapHuman to accomplish our target. Our CapHuman is built upon the recent pre-trained text-to-image diffusion model, Stable Diffusion~\cite{rombach2022SD}, which serves as a general representative vision generator. As a basis, we aim to unlock its potential for generalizable identity preservation and fine-grained head control. Instead of test-time fine-tuning the pre-trained model, we embrace the ``encode then learn to align" paradigm, which guarantees generalizable identity preservation for new individuals without cumbersome tuning at inference. Specifically, our CapHuman encodes the global and local identity features and then aligns them into the latent feature space. Additionally, our generative model is equipped with fine-grained head control by leveraging the 3D Morphable Face Model~\cite{FLAME, yang2021multiple, zhou2024headstudio}. It provides a flexible and 3D-consistent way to control the head via the parameter tuning, once we build the 3D facial representation to the reference image correspondence. With the 3D-aware facial prior, the local geometric details are better preserved.

We introduce HumanIPHC, a new challenging and comprehensive benchmark for identity preservation, text-to-image alignment, and head control precision evaluation. Our CapHuman achieves impressive qualitative and quantitative results compared with other established baselines, demonstrating the effectiveness of our proposed method. 

Overall, our contributions can be summarized as follows:
\begin{itemize}
    \item We propose a novel human-centric image synthesis task that generates specific individual portraits with various head positions, poses, facial expressions, and illuminations in different contexts given one reference image.
    \item We propose a new framework CapHuman. We embrace the ``encode then learn to align" paradigm for generalizable identity preservation without tuning at inference, and introduce 3D facial representation to provide fine-grained head control in a flexible and 3D-consistent manner.
    \item To the best of our knowledge, our CapHuman is the first framework to preserve individual identity while enabling text and head control in human-centric image synthesis.
    \item We introduce a new benchmark HumanIPHC to evaluate identity preservation, text-to-image alignment, and head control ability. Our method outperforms other baselines.
\end{itemize}
\section{Related Work}
\label{sec:related_work}
\subsection{Text-to-Image Synthesis}
There has been significant advancement in the field of text-to-image synthesis. With the emergence of large-scale data collections such as LAION-5B~\cite{schuhmann2022laion} and the support of powerful computation resources, large generative models bloom in abundance. One of the pathways is driven by diffusion models. Diffusion models~\cite{ho2020ddpm} are easily scalable without instability and mode collapse of adversarial training~\cite{goodfellow2020GAN}. They have achieved amazing results in generating photo-realistic and content-rich images with high fidelity. Imagen~\cite{saharia2022imagen}, GLIDE~\cite{nichol2021glide}, and DALL-E 2~\cite{ramesh2022dalle2} directly operate the denoising process in the pixel space. Instead, Stable Diffusion~\cite{rombach2022SD} performs it in the latent space to enable training under the limited resources scenarios while retaining the capability of high-quality image generation. Besides, some works research on auto-regressive modeling~\cite{yu2022parti} or masked generative modeling~\cite{chang2023muse}. Recently, GigaGAN~\cite{kang2023giga} has explored the potential of the traditional GAN framework~\cite{karras2019style} for large-scale training on the same large datasets and can synthesize high-resolution images as well.

\begin{figure*}[t!]
\centering
\includegraphics[width=0.92\textwidth]{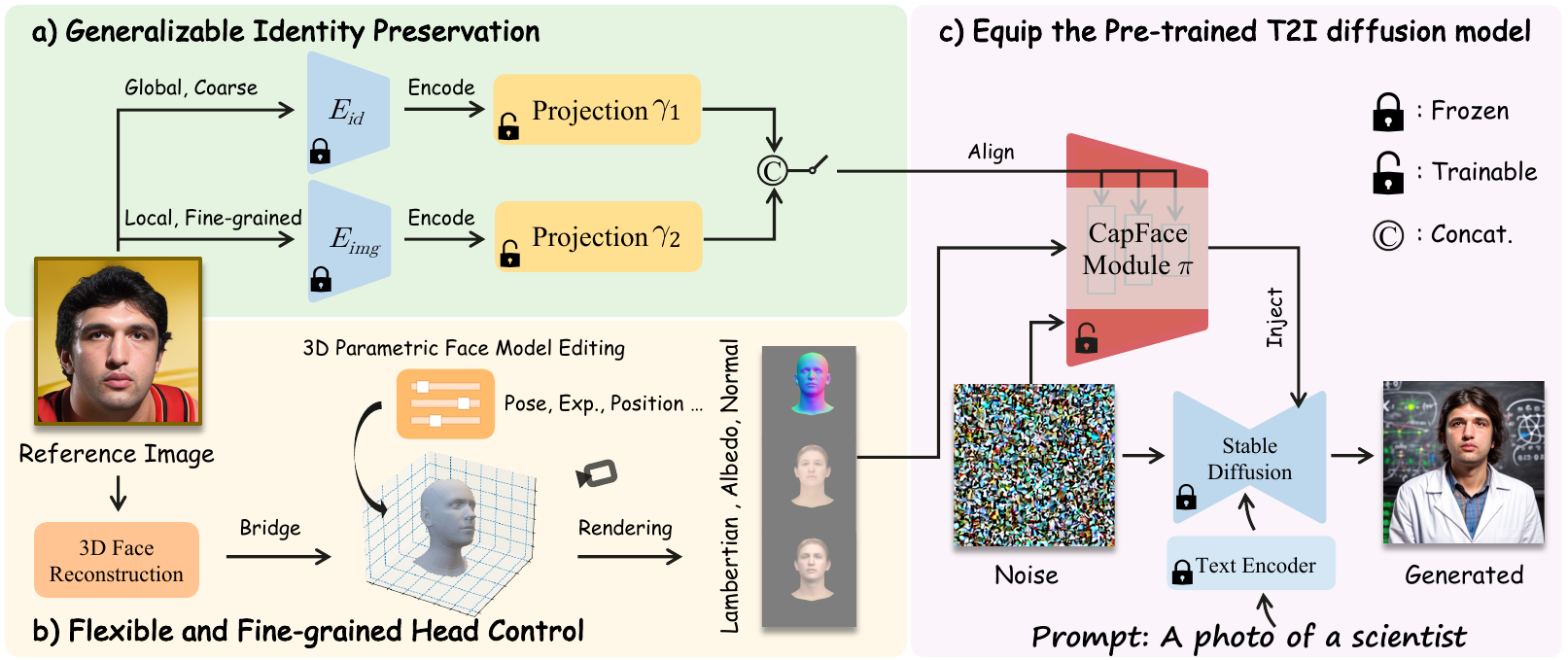}
\vspace{-0.5em}
\caption{
\textbf{Overview of CapHuman.} Our CapHuman stands upon the pre-trained T2I diffusion model. a) We embrace the ``encode then learn to align" paradigm for generalizable identity preservation. b) The introduction of the 3D parametric face model enables flexible and fine-grained head control. c) We learn a CapFace module $\pi$ to equip the pre-trained T2I diffusion model with the above capabilities.
}
\vspace{-1.8em}
\label{fig:framework}
\end{figure*}

\subsection{Personalized Image Generation}
Given a small subset of reference images, the personalization for text-to-image diffusion models aims to endow the pre-trained models with the capability of preserving the identity of a specific subject. Although large text-to-image diffusion models have learned strong semantic priors, they are still lacking the ability of identity preservation. A series of approaches are proposed to compensate for this missing ability by fine-tuning the pre-trained models. Textual Inversion~\cite{gal2023TI} introduces a new word embedding for the user-provided concept. However, too few parameters limit the expressiveness of the output space. DreamBooth~\cite{ruiz2023dreambooth} fine-tunes the entire UNet backbone with a unique identifier. A class-specific prior preservation loss is further used to overcome the overfitting problem, due to the limited number of reference images. Considering the efficiency of fine-tuning, LoRA~\cite{hu2021lora} only learns the residual of the model with low-rank matrices. These methods follow the ``test-time fine-tuning" paradigm and need to personalize the pre-trained model for each subject. As a result, all of them fall short of fast and generalizable personalization. To address the aforementioned problem, a few works~\cite{shi2023instantbooth, jia2023taming, xiao2023fastcomposer} pursue a tuning-free method. The main idea is to learn a generalizable encoder for the novel subject and preserve the text control, free from additional fine-tuning at test time.

\subsection{Controllable Human Image Generation}
Text-conditioned methods~\cite{huang2023avatarfusion,shen2023controllable,xu2023seeavatar,gan2023efficient} have shown remarkable capability in human/avatar generation. The text condition is awesome, but still unsatisfactory for real-world applications like human image generation, which requires more fine-grained control. The challenge is how to structurally control the existing pre-trained text-to-image models. ControlNet~\cite{zhang2023controlnet} and T2I Adapter~\cite{mou2023t2iadapter} design an adapter to align the new and external control signal with the original internal representation of the pre-trained text-to-image models. They both provide pose-guided conditional generation but fail to preserve the identity. In addition, DiffusionRig~\cite{ding2023diffusionrig} supports personalized facial editing with head control. The proposed framework cannot provide the text editing ability, limiting its generative capability.

\section{Method}
\setlength{\abovedisplayskip}{8pt}
\setlength{\belowdisplayskip}{7pt}
\label{sec:method}
\subsection{Preliminary}
\paragraph{Stable Diffusion~\cite{rombach2022SD}} is a popular open-source text-to-image generation framework, that achieves great progress in high-resolution and content-rich image generation. It has attracted considerable interest and is applied in several tasks~\cite{Liu_2023_zero123, xu2023dream3d, blattmann2023align, zhang2023sifu, yang2024doraemongpt, zhou2024migc, quan2024psychometry}. Stable Diffusion belongs to the family of the latent diffusion models. By compressing the data into the latent space, it enables more efficient scalable model training and image generation. This framework is composed of two stages. First, it trains an autoencoder $\mathcal{E}$ to map the original image $x$ into the lower-dimensional latent representation $z=\mathcal{E}(x)$. Then, in the latent space, a time-conditional UNet denoiser predicts the added noise at different timesteps. For the text condition, this model employs the cross-attention mechanism~\cite{vaswani2017attention} to understand the semantics of text prompts. Put it together, the denoising objective can be formulated as follows:
\begin{align}
    \mathcal{L}_{LDM} = E_{z,c,\epsilon \sim \mathcal{N}(\boldsymbol{0}, \mathbf{I}),t \sim \mathcal{U}(1, T)} \left[ \left\| \epsilon_\theta(z_t,t,c) - \epsilon \right\|_2 \right],
\end{align}
where $z_t$ is the noisy latent code, $c$ is the text embedding, $\epsilon$ is sampled from the standard Gaussian distribution, and $t$ is the timestep. Pre-trained on large-scale internet data, Stable Diffusion has learned strong semantic and relation priors for natural and high-quality image generation.

\vspace{-1.3em}
\paragraph{FLAME~\cite{FLAME}} is one of the expressive 3D Morphable Models~(3DMM)~\cite{FLAME, paysan2009bfm, booth20163d, booth2018large, blanz2023morphable}. It is a statistical parametric face model that captures variations in shape, pose, and facial expression. Given the coefficients of shape $\beta$, pose $\theta$, and expression $\psi$, the model can be described as:
\begin{align}
    M(\beta, \theta, \psi) = W(T_P(\beta, \theta, \psi), J(\beta), \theta, \mathcal{W}),
\end{align}
where $T_P$ is rotated around joints $J$ linearly smoothed by blendweight $\mathcal{W}$. Here, $T_P$ denotes the template with added shape, pose, and expression offsets. In other words, it is flexible for us to control the facial geometry by adjusting or tuning the parameters of $\beta$, $\theta$, and $\psi$ within a range.

\subsection{Overview}
In this work, we consider a novel human-centric image synthesis task. Given only one reference face image $I$ indicating the individual identity, our goal is to generate photo-realistic and diverse images for the specific identity with different head positions, poses, facial expressions, and illuminations in different contexts, driven by the text prompt $\mathcal{P}$ and the head condition $\mathcal{H}$. Input as a triplet data pair $(I, \mathcal{P}, \mathcal{H})$, we learn a model $\mathcal{G}$ as our generative model to produce a new image $\hat{I}$. The pipeline can be defined as:
\begin{align}
    \hat{I} = \mathcal{G}(I, \mathcal{P}, \mathcal{H}).
\end{align}
To accomplish this task, ideally, the model $\mathcal{G}$ should be equipped with the following functionalities: (1) basic object and human image generation capability. (2) generalizable identity preservation ability. (3) flexible and fine-grained head control. Recently, large pre-trained text-to-image diffusion models~\cite{rombach2022SD, saharia2022imagen, ramesh2022dalle2} have shown incredible and impressive generative ability. They are born with the implicit knowledge of our world and human society, which serves as a good starting point for our consolidation. We propose a new framework CapHuman, which is built upon the pre-trained text-to-image diffusion model, Stable Diffusion~\cite{rombach2022SD}. Although Stable Diffusion has the in-born generation capability, it still lacks the ability of identity preservation and head control, limiting its application in our scenario. We aim to endow the pre-trained model with the above two abilities by introducing a CapFace module $\pi$. Our pipeline exhibits several advantages: \textit{\underline{well-generalizable}} identity preservation that needs no time-consuming fine-tuning for each new individual, \textit{\underline{3D-consistent}} head control that incorporates 3DMM to support fine-grained control, and \textit{\underline{plug-and-play}} property that is compatible with rich off-the-shelf base models. \S~\ref{sec:id_preservation} introduces the generalizable identity preservation module. \S~\ref{sec:head_control} concentrates on the flexible and fine-grained head control capability. \S~\ref{sec:training_inference} presents the training and inference process. The overall framework is shown in Figure~\ref{fig:framework}.
\vspace{-0.7em}
\subsection{Generalizable Identity Preservation}
\label{sec:id_preservation}
The most straightforward solution~\cite{gal2023TI, ruiz2023dreambooth, hu2021lora} is to fine-tune the pre-trained model with the given reference image. Though the model can preserve the identity in this case, it sacrifices the generality. The fine-tuning process forces the model to memorize the specific individual. When a new individual comes, it needs to re-train the model, which is cumbersome. Instead, we advocate the ``encode then learn to align" paradigm, that is, we treat identity preservation as one of the generalizable capabilities that our model should have. We formulate it as a learning task. The task requires our model to learn to extract the identity information from one reference image and preserve the individual identity in the image generation. We break it down into two steps. 
\vspace{-1.9em}
\paragraph{Encode global and local identity features.} In the first step, the reference face image $I$ is encoded into identity features at different granularities. Here, we consider two types of identity features: (1) \textbf{global coarse feature} represents the key and typical characteristics of the human face. We use the feature extractor $E_{id}$ pre-trained on the face recognition task~\cite{schroff2015facenet} to obtain the global face embedding $\mathbf{f}_{global} = E_{id}(I) \in \mathbb{R}^{1\times d_1}$. The global feature captures the key information to help distinguish it from other identities, but some appearance details might be overlooked. (2) \textbf{local fine-grained feature} depicts more facial details, which can further enhance the fidelity of face image generation. We leverage the CLIP~\cite{radford2021clip} image encoder $E_{img}$ to extract local patch image feature $\mathbf{f}_{local} = E_{img}(I) \in \mathbb{R}^{N \times d_2}$. Note that we only keep the face area by segmentation~\cite{OMGSeg,li2023transformer} and the irrelevant background is removed.

\vspace{-1.9em}
\paragraph{Learn to align into the latent space.} In the second step, our model $\pi$ learns to align the identity features into its feature space. As identity features contain high-level semantic information, we inject them like Stable Diffusion~\cite{rombach2022SD} treats the text. We embed the global and local features into the latent identity feature $\mathbf{f}_{id}$:
\begin{align}
    \mathbf{f}_{id} = [\gamma_1(\mathbf{f}_{global}); \gamma_2(\mathbf{f}_{local})] \in \mathbb{R}^{(1+N)\times d},
\end{align}
where $\gamma_1$, $\gamma_2$ are projection layers and $[;]$ denotes the concatenation operation.
Then, the latent identity feature is processed by the cross-attention mechanism~\cite{vaswani2017attention}, attending to the latent feature $\mathbf{f}_l$ in $\pi$, as formulated in the following way:
\begin{align}
    \text{Attention}(Q, K, V) = \text{softmax}(\frac{QK^T}{\sqrt{d_k}})V,
\end{align}
where the query, key and value are defined as $Q=\phi_Q(\mathbf{f}_{l})$, $K=\phi_K(\mathbf{f}_{id})$, $V=\phi_V(\mathbf{f}_{id})$. And $\phi_Q$, $\phi_K$, $\phi_V$ are linear projections. By inserting the identity features into the latent feature space in the denoising process, our model can preserve the individual identity in the image synthesis. The combination of global and local features not only strengthens the recognition of individual identity but also complements the facial details in the human image generation. The ``encode then learn to align" paradigm guarantees our model is generalizable for new individuals without the need for extra tuning in the inference time. 

\subsection{Flexible and Fine-grained Head Control}
\label{sec:head_control}
Human-centric image generation favors flexible, fine-grained, and precise control over the human head. It is desirable to have the ability to put the head everywhere in any pose and expression in the human image synthesis. However, the powerful pre-trained text-to-image diffusion model lacks this control. It is believed that the pre-trained model has learned internal structural priors regarding the generation of diverse human images with varying head positions, poses, facial expressions, and illuminations. We aim to unlock its capability by introducing an appropriate control signal as a trigger. The first question is: what constitutes a good representation for this signal? 

\vspace{-1.2em}
\paragraph{Bridge 3D facial representation.} We pay attention to the popular 3DMM FLAME~\cite{FLAME}. It constructs a compact latent space to represent the shape, pose, and facial expression separately. It provides a friendly and flexible interface to edit the facial geometry, \eg changing the head pose, and facial expression with varied parameters. In our setting, we bridge the input reference image $I$ and the 3D facial representation. We use DECA~\cite{DECA} to reconstruct the specific 3D head model with detailed facial geometry from a single image. Then, we transform it into a set of pixel-aligned condition images including Surface Normal, Albedo, and Lambertian rendering. They contain the position, local geometry, albedo, and illumination information~\cite{ding2023diffusionrig}. 

\vspace{-1.5em}
\paragraph{Equip with 3D-consistent head control.} We attempt to equip the pre-trained generative model with the ability to respond to the control signal. Given the head condition $\mathcal{H}=\{I_{Normal}, I_{Albedo}, I_{Lambertian}\}$, we obtain the feature map $\mathcal{F}_t$. The process is defined as:
\begin{align}
    \mathcal{F}_t = \pi(z_t, t, \mathcal{H}, \mathbf{f}_{id}).
\end{align}
Because the head condition images are coarse facial appearance representations, we incorporate the identity features to strengthen the local details. In order to force the CapFace module $\pi$ to focus on the facial area, we predict the facial mask $\mathcal{M}$ from the head condition $\mathcal{H}$. Finally, the masked feature map $\mathcal{F}_t \odot \mathcal{M}$ is injected into the original feature space of the pre-trained model. Considering the low-level characteristics of head control and plug-and-play property, we adopt the side network design like ControlNet~\cite{zhang2023controlnet}. CapFace module $\pi$ shares a similar structure with the Stable Diffusion encoder. The feature map is element-wise aligned with that in the decoder part of Stable Diffusion for each layer. By embedding the new control signal, the pre-trained model is endowed with the ability of head control. The introduction of the 3D parametric face model enables 3D-consistent control of the human head. 
\setlength{\abovedisplayskip}{10pt}
\setlength{\belowdisplayskip}{10pt}
\subsection{Training and Inference}
\label{sec:training_inference}
\paragraph{Training objective.} We calculate the denoising loss between the predicted and groundtruth noise, with the mask prediction loss. The training objective for the model optimization is formulated as:
\begin{align}
    \mathcal{L} = & \left\| \epsilon_\theta(z_t,t,c, \pi(z_t, t, \mathcal{H}, \mathbf{f}_{id})) - \epsilon \right\|_2 + \lambda \left\| \mathcal{M} - \mathcal{M}_{gt} \right\|_2, 
\end{align}
where $\mathcal{M}_{gt}$ is the groundtruth facial mask, and we set $\lambda=1$. We keep $\epsilon_{\theta}$ frozen and train the CapFace module $\pi$.

\vspace{-1.3em}
\paragraph{Time-dependent ID dropout.} Our model might focus more on the identity features due to the entanglement of the head pose information in the reference image, which results in weak control of the head condition. Inspired by the fact that the denoising process in the diffusion model is progressive and the appearance is concentrated at the later stage~\cite{ho2020ddpm}, we propose a time-dependent ID dropout regularization strategy that discards the identity feature at the early stage to alleviate the issue. We formulate the strategy in the following:
\begin{align}
    \mathcal{F}_t^{\dag} = \begin{cases} \pi(z_t, t, \mathcal{H}, \mathbf{f}_{id}), & t < \tau, \\
    \pi(z_t, t, \mathcal{H}, \varnothing), & \text{otherwise}, \\
    \end{cases}
\end{align}
where $t$ is the timestep in the diffusion process, $\tau$ is the start timestep, and $\mathcal{F}_t^{\dag}$ is the feature map.

\vspace{-1.3em}
\paragraph{Post-hoc Head Control Enhancement.} To enhance the head control of our generative model, we optionally fuse the feature map with others from the head control model $\pi^{\star}$ at inference:
\begin{align}
\label{eq:fusion}
    \mathcal{F}_t^{\ddag} = \pi(z_t, t, \mathcal{H}, \mathbf{f}_{id}) + \alpha \cdot \pi^{\star}(z_t, t, \mathcal{H}, \varnothing),
\end{align}
where $\alpha$ is the control scale and $\mathcal{F}_t^{\ddag}$ is the feature map.
\begin{figure*}[t!]
\centering
\includegraphics[width=0.95\textwidth]{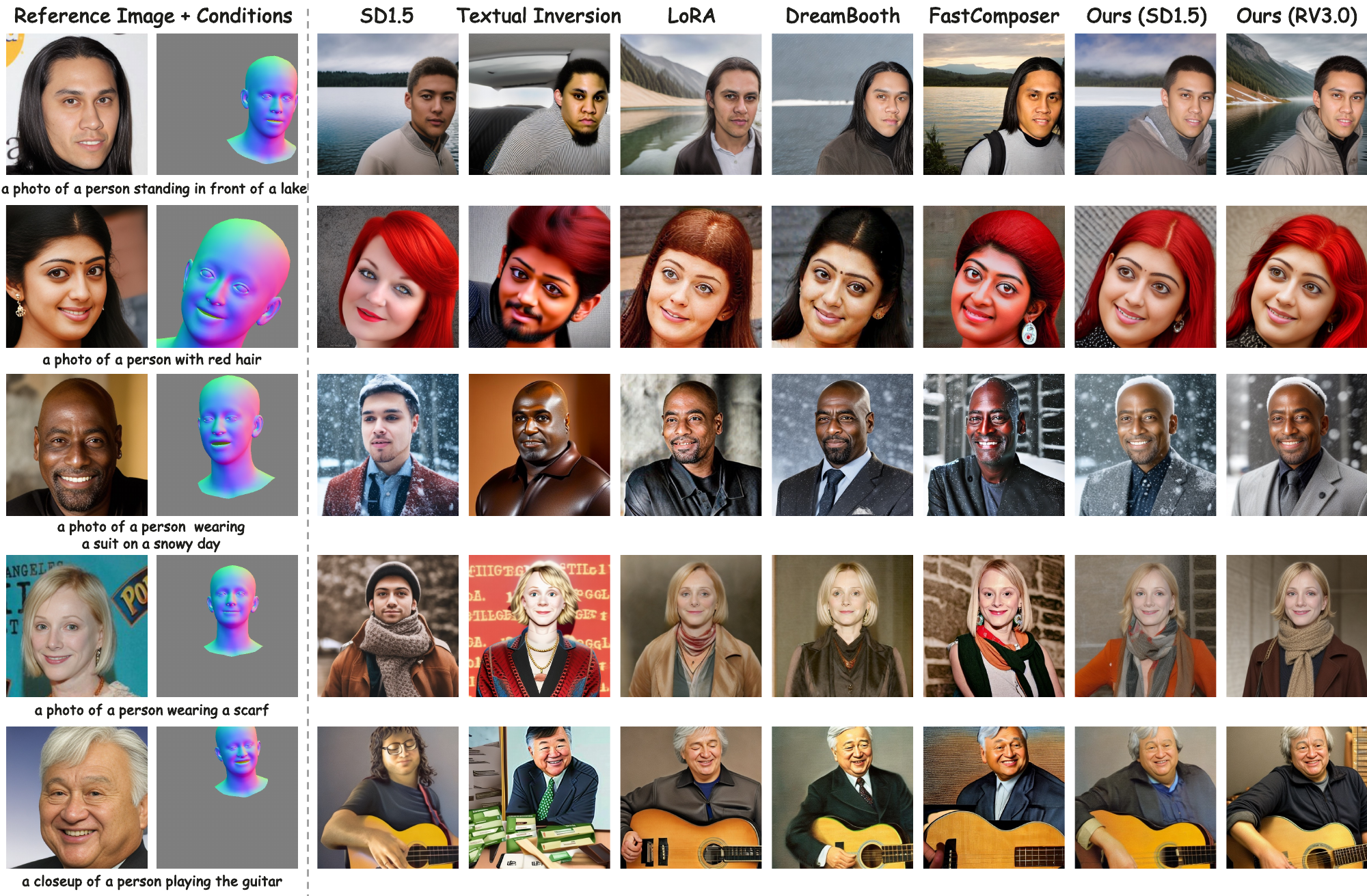}
\vspace{-1.0em}
\caption{
\textbf{Qualitative results.} Our CapHuman can produce identity-preserved, photo-realistic portraits with various head positions and poses in different contexts. Also, our model can be combined with the pre-trained model, \eg RealisticVision~\cite{Realistic} in the community flexibly.
}
\vspace{-1.0em}
\label{fig:main_results}
\end{figure*}
\section{Experiments}
\label{sec:exp}
\subsection{Training setup}
We train our model on CelebA~\cite{liu2015celeba}, which is a large-scale face dataset with more than 200K celebrity images, covering diverse pose variations. For data preprocessing, we crop and resize the image to the size of 512 $\times$ 512 resolution. Following~\cite{karras2019style}, we crop and align the face region for the reference image. We use BLIP~\cite{li2022blip} for image captioning. We choose ViT-L/14 as the CLIP~\cite{radford2021clip} image encoder. Our model is based on Stable Diffusion V1.5~\cite{rombach2022SD}. The learning rate is 0.0001 and the batch size is 128. We use AdamW~\cite{loshchilov2017adamw} for the optimization.

\subsection{Qualitative Analysis}
\paragraph{Visual comparisons.}We focus on the one-shot setting where only one reference image is given. We compare our method with the established techniques including Textual Inversion~\cite{gal2023TI}, DreamBooth~\cite{ruiz2023dreambooth}, LoRA~\cite{hu2021lora} and FastComposer~\cite{xiao2023fastcomposer}. These methods are designed for personalization and lack of head control. For fair comparisons, we combine them with ControlNet~\cite{zhang2023controlnet}, since ControlNet can provide facial landmark-driven control. Also, landmark-guided ControlNet~\cite{zhang2023controlnet} is one of our baselines. The visual qualitative results are presented in Figure~\ref{fig:main_results}. Obviously, landmark-guided ControlNet cannot preserve the individual identity. The fine-tuning baselines can preserve the individual identity to a certain extent. However, they suffer from the overfitting issue. The input prompt might not take effect in some cases. It suggests that these methods sacrifice the diversity for the identity memorization. Compared with the state-of-the-art approaches, our method shows competitive and impressive generative results with good identity preservation. Given only one reference photo, our CapHuman can produce photo-realistic and well-identity-preserved images with various head positions and poses in different contexts. 
\begin{figure}[t!]
\centering
\includegraphics[width=1.0\linewidth]{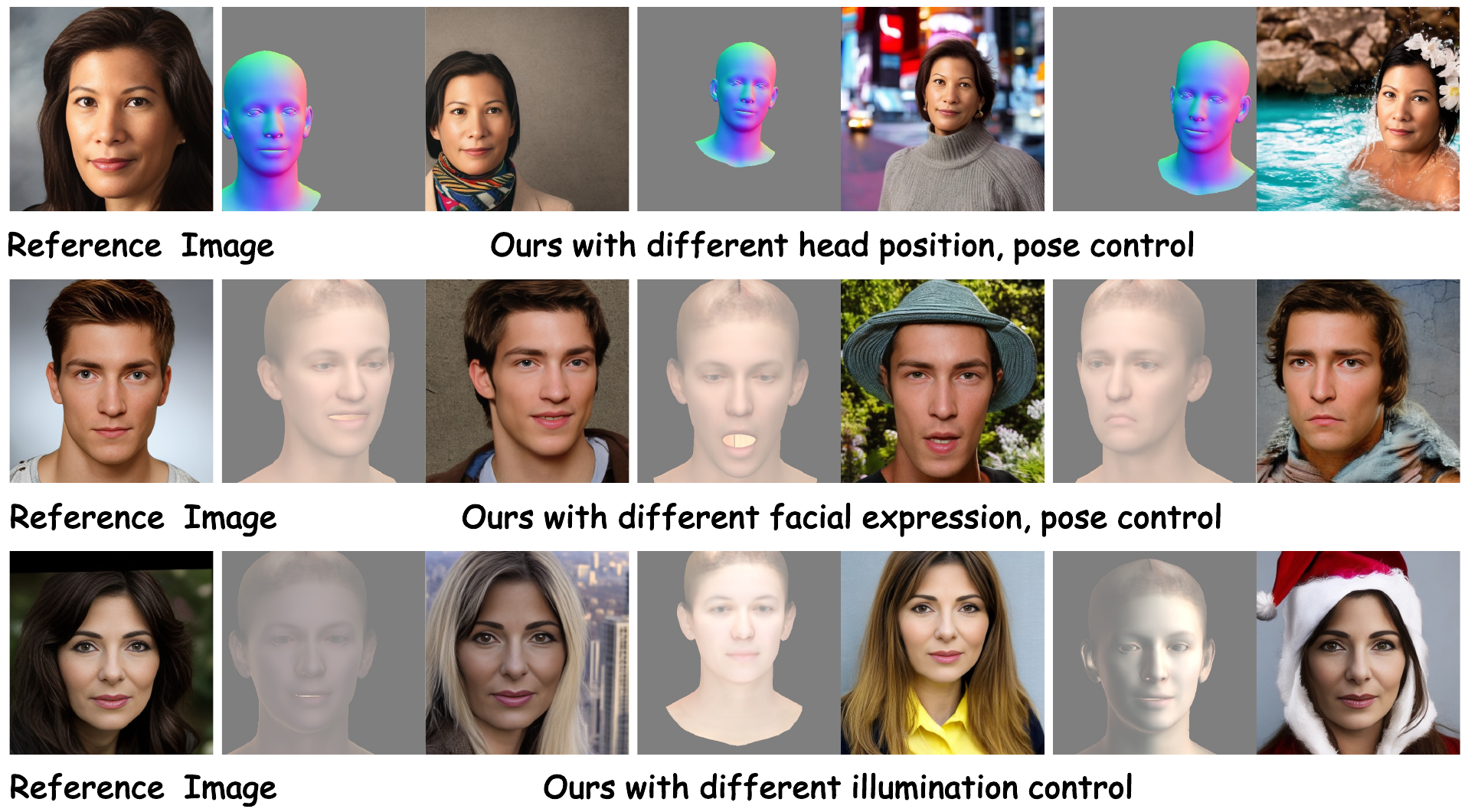}
\vspace{-1.9em}
\caption{
\textbf{Head position, pose, facial expression, and illumination control.} Our method offers the 3D-consistent head control.
}
\vspace{-1.0em}
\label{fig:position_exp_control}
\end{figure}
\begin{figure}[t!]
\centering
\includegraphics[width=1.0\linewidth]{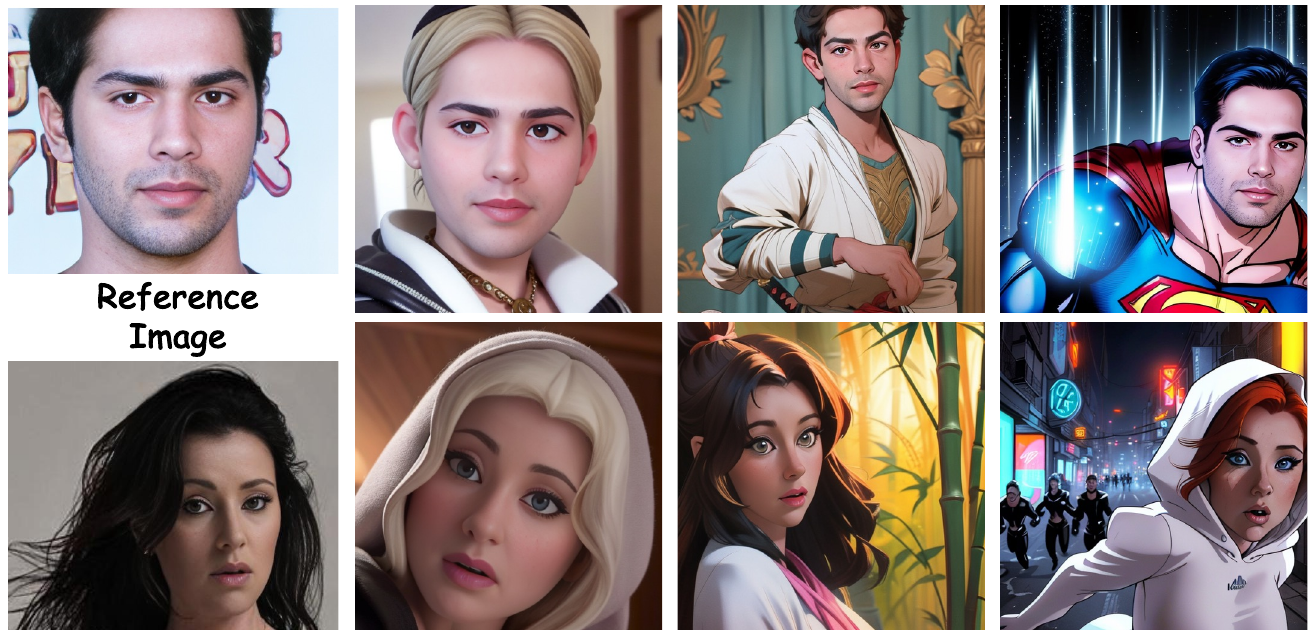}
\vspace{-1.9em}
\caption{
\textbf{Adapt our model to other pre-trained models.}
Our model can be adapted to generate portraits in different styles.
}
\vspace{-1.4em}
\label{fig:our_pretrain}
\end{figure}
\vspace{-2.4em}
\paragraph{Head control capability.} Figure~\ref{fig:position_exp_control} shows the head control capability of our CapHuman. The results demonstrate our CapHuman can offer 3D-consistent control over the human head in diverse positions, poses, facial expressions, and illuminations. More results can be found in the appendix.
\vspace{-1.4em}
\paragraph{Adapt to other pre-trained models.} The plug-and-play property enables our model can be adapted to other pre-trained models~\cite{toonyou, disney-pixar-cartoon,comic-babes} in the community seamlessly. The results are presented in Figure~\ref{fig:our_pretrain}. More visual results with more styles can be found in the appendix.

\subsection{Quantitative Analysis}
\paragraph{Benchmark.} We introduce a new challenging and comprehensive benchmark HumanIPHC for identity preservation, text-to-image alignment, and head control precision evaluation. We select 100 identities from the CelebA~\cite{liu2015celeba} test split. They consist of different ages, genders, and races. We collect 35 diverse prompts and 10 different head conditions with various positions and poses. Three different images are generated for each combination.
\vspace{-1.3em}
\paragraph{Evaluation metrics.} We evaluate the effectiveness of our proposed method in the following three dimensions: (1) Identity Preservation. We apply a face recognition network~\cite{schroff2015facenet} to extract the facial identity feature from the face region. The cosine similarity between the reference image and the generated image is used to measure the facial identity similarity. (2) Text-to-Image Alignment. We use the CLIP score as the metric. The CLIP~\cite{radford2021clip} score is calculated as the pairwise cosine similarity between the image and text features. In addition, we report the prompt accuracy. It is the classification accuracy between the generated image and a set of candidate prompts. We check whether the prompt with the largest CLIP score is the prompt used to generate or not.  (3) Head Control Precision. We compute the root mean squared error~(RMSE) between the DECA~\cite{DECA} code estimated from the generated image and the given condition. We divide the DECA code into four groups: Shape, Pose, Expression, and Lighting.
\vspace{-1.3em}
\paragraph{Quantitative results.} Table~\ref{tab:main_tb} shows the evaluation results on our benchmark. For identity preservation, Textual Inversion~\cite{gal2023TI}, LoRA~\cite{hu2021lora}, and DreamBooth~\cite{ruiz2023dreambooth} can improve the performance on identity similarity. Their abilities depend on the scale of the trainable parameters. DreamBooth fine-tunes the entire backbone while Textual Inversion only trains the word embedding. As a result, DreamBooth shows better results. By learning to encode the identity information, our model achieves generalizable identity preservation capability, surpassing DreamBooth~\cite{ruiz2023dreambooth} and FastComposer~\cite{xiao2023fastcomposer} by $15\%$ and $21\%$, respectively. For text-to-image alignment, the fine-tuning methods fall into the overfitting problem under the one-shot setting. They sacrifice prompt diversity for better identity preservation. In contrast, our method can still maintain a high level of prompt control. For head control precision, our method shows remarkable improvement in Shape, Expression, and Lighting metrics, \ie, $5\%$, $7\%$, $7\%$ compared with the second best results. We attribute this to the introduction of the 3D facial prior.
% \vspace{-2.2em}
\begin{table}[t]
% \small
\centering
\setlength\tabcolsep{3pt}
\resizebox{1.0\linewidth}{!}{
\begin{tabular}{l|cc|cc|cccc}
\hline
\rowcolor[gray]{0.9}
& \multicolumn{2}{c|}{Identity Preservation} & \multicolumn{2}{c|}{Text-to-Image Alignment} & \multicolumn{4}{c}{Head Control Precision} \\ 
\rowcolor[gray]{0.9}
Method &  Generalizable &$\uparrow$ ID sim.   & $\uparrow$  CLIP score  & $\uparrow$  Prompt acc. & $\downarrow$ Shape   & $\downarrow$  Pose  & $\downarrow$  Exp.& $\downarrow$ Light. \\ \hline \hline 
ControlNet~\cite{zhang2023controlnet} & \xmark   & 0.0534  & \textbf{0.2479} & \textbf{90.32\%} & 0.2722  & 0.0494 & 0.3584 & 0.2718  \\
Textual Inversion~\cite{gal2023TI}  & \xmark & 0.4857  &  0.1561 &  13.70\% & 0.2075 & 0.0516 & 0.2530 & 0.2579\\ 
LoRA~\cite{hu2021lora} & \xmark   & 0.5860  & 0.1897 & 35.96\% & 0.1648 & 0.0446 & 0.2039 & 0.1634  \\
DreamBooth~\cite{ruiz2023dreambooth} & \xmark & 0.6860 & 0.1873 & 39.21\% & 0.1542 & 0.0441 & 0.1922 & 0.1729\\
FastComposer~\cite{xiao2023fastcomposer} & \cmark & 0.6191 & 0.2150 & 68.52\% & 0.1851 & 0.0611 & 0.2119 & 0.1861\\ \hline
Ours & \cmark & \textbf{0.8363} & 0.2256 & 74.17\% & \textbf{0.1020} & \textbf{0.0436} & \textbf{0.1241} & \textbf{0.0965}\\
\hline
\end{tabular}
}
\vspace{-0.8em}
 \caption{
     \textbf{Comparisons with the established state-of-the-art methods.} Our CapHuman outperforms other baselines for better identity preservation and better head control. Compared with other personalization methods, our method can still keep a high level of prompt control. \textbf{Bold} denotes the best result. }
    \label{tab:main_tb}
\vspace{-1.0em}
\end{table}

\begin{table}[t]
\centering
%#################################################
% Ablation on global and local identity features
%#################################################
\begin{minipage}{0.29\linewidth}{
\centering
\small
\setlength\tabcolsep{3pt}
\begin{tabular}{l|c}
\rowcolor[gray]{.9}
\hline 
Method &  $\uparrow$ ID sim.  \\ \hline \hline 
w/o global \& local feat.    &  0.3915 \\
w/o local feat.   & 0.7725   \\
w/o global feat.   &  0.8095 \\ 
w/ global \& local feat. &  \textbf{0.8429} \\
\hline
\end{tabular}
\captionsetup{width=1.8\linewidth}
\vspace{-1.2em}
\captionof{table}{\textbf{Ablation on ID features.}}
\label{tab:ab_feat}
}
\end{minipage}
% \hspace{2em}
\hspace{7em}
%#################################################
% Ablation on the effect of N in the local feature
%#################################################
\begin{minipage}{0.29\linewidth}{
\centering
\small
\setlength\tabcolsep{3pt}
\begin{tabular}{l|c}
\rowcolor[gray]{.9}
\hline 
Num. $N$ &  $\uparrow$ ID sim.  \\ \hline \hline 
% $32$    &    XXXX\\
% $64$    &    0.7990\\
% $128$   &   0.7971\\ 
% $257$   &  \textbf{0.8179}  \\
$32$    &    0.8370\\
$64$    &    0.8376\\
$128$   &   0.8182\\ 
$257$   &  \textbf{0.8429}  \\
\hline
\end{tabular}
\captionsetup{width=1.2\linewidth}
\vspace{-1.2em}
\captionof{table}{\textbf{Effect of $N$.}}
\label{tab:ab_ntokens}
}\end{minipage}
\vspace{-1.0em}
% \hspace{5em}
\end{table}
%##################################################################################################

\begin{figure}[t!]
\centering
\includegraphics[width=1.0\linewidth]{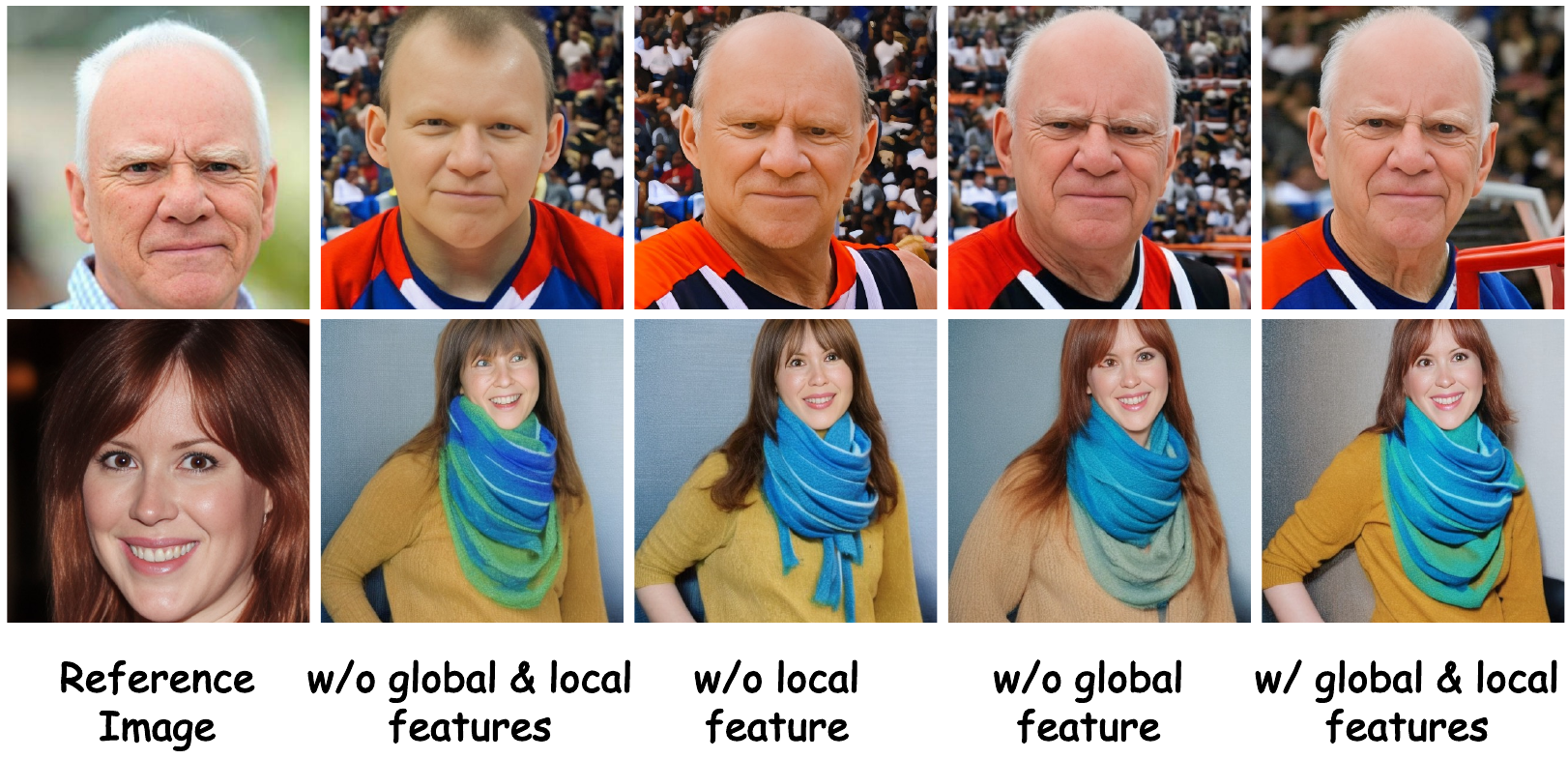}
\vspace{-1.7em}
\caption{
\textbf{Visual results of global and local identity features.} Both global and local features contribute to identity preservation.
}
\vspace{-2.0em}
\label{fig:ab_id}
\end{figure}
\subsection{Ablation Studies}
We perform the ablation studies on a small subset with 10 identities to study the effectiveness of our design.
\vspace{-1.3em}
\paragraph{Effect of global and local identity features.} We investigate the importance of global and local features for identity preservation. In Table~\ref{tab:ab_feat}, we present the identity similarity comparison. As expected, both global and local identity features contribute to identity preservation. The performance drops when removing the global or local feature individually. Furthermore, we illustrate the effectiveness of the identity features in Figure~\ref{fig:ab_id}. We can observe that our model cannot preserve the individual identity if no identity features are involved during the image generation. With the global identity feature, we can recognize the identity basically. Additionally, the local feature complements the details and enhances the facial fidelity.
\vspace{-1.4em}

\paragraph{Effect of the number $N$ in the local identity feature.} We study the effect of the number $N$ in the local identity feature. As reported in Table~\ref{tab:ab_ntokens}, we find the compression of the local identity feature can hurt the performance of identity preservation. It is better to make full use of the local identity features in human face image generation.
\begin{table}[t]
\small
\centering
\setlength\tabcolsep{6.5pt}
\resizebox{1.0\linewidth}{!}{
\begin{tabular}{l|cccc}
\rowcolor[gray]{.9}
\hline 
Method &  $\downarrow$ Shape   & $\downarrow$  Pose  & $\downarrow$  Exp. & $\downarrow$ Light. \\ \hline \hline 
w/o 3DMM & 0.2909 & 0.0501 & 0.3967 & 0.2899\\ 
w/ 3DMM~(Ours) & \textbf{0.1381} & \textbf{0.0262} & \textbf{0.1639} & \textbf{0.1196} \\
\hline
\end{tabular}
}
\vspace{-0.8em}
 \caption{
     \textbf{Ablation on 3DMM.} Ours with 3DMM achieves significant improvement in head control precision.}
    \label{tab:ab_controlnet}
\vspace{-1.5em}
\end{table}
\begin{figure}[t!]
\centering
\includegraphics[width=1.0\linewidth]{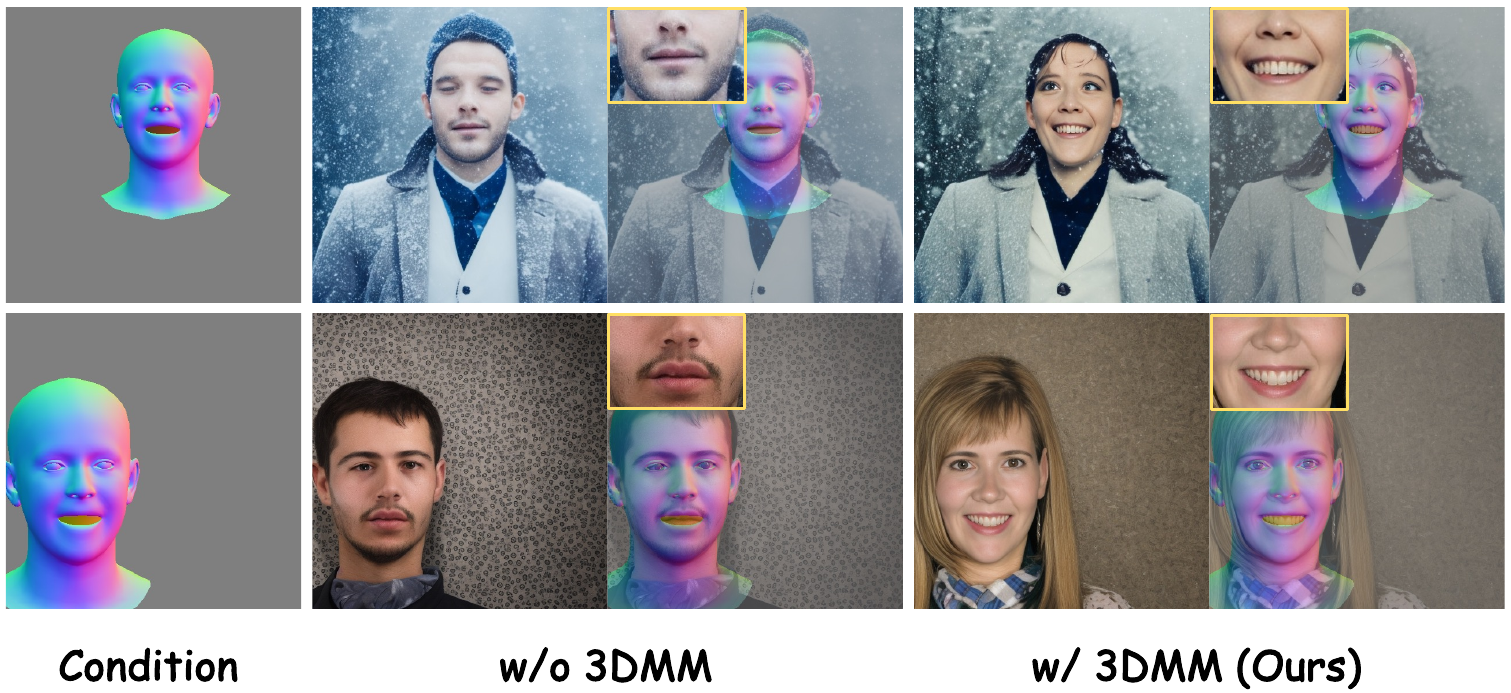}
\vspace{-2.2em}
\caption{
\textbf{Visual comparison on 3DMM.} Ours with 3DMM shows more fine-grained control results with local details.
}
\vspace{-2.0em}
\label{fig:ab_pose}
\end{figure}
\vspace{-1.4em}
\paragraph{Ablation on 3DMM.} 
We validate the effectiveness of 3DMM. We remove the identity preservation module. Table~\ref{tab:ab_controlnet} shows the results. With 3DMM, our method shows significant improvement in head control precision. The introduction of the 3D facial representation brings more information such as local geometry and illumination. Figure~\ref{fig:ab_pose} confirms the more precise head control of our method.

\vspace{-1.4em}
\paragraph{Influence of the ID dropout start timestep $\tau$.} We study the influence of the ID dropout start timestep $\tau$. As shown in Table~\ref{tab:ab_tau}, with more time identity features participate in the denoising process, our model shows stronger identity preservation capability. However, the pose metric gets worse. In the learning process, our model might concentrate more on the identity feature and overlook the pose condition. The experimental results prove that the time-dependent ID dropout strategy plays a role in the tradeoff between identity preservation and head pose control.
\begin{table}[t]
\small
\centering
\setlength\tabcolsep{5pt}
\resizebox{1.0\linewidth}{!}{
\begin{tabular}{l|c|cccc}
\rowcolor[gray]{.9}
\hline 
Method & $\uparrow$ ID sim.& $\downarrow$ Shape   & $\downarrow$  Pose  & $\downarrow$  Exp. & $\downarrow$ Light. \\ \hline \hline 
$\tau=0$ & 0.3915 & 0.1381 & \textbf{0.0262} & 0.1639 & 0.1196 \\
$\tau=300$ & 0.6600 & 0.1257 & 0.0292 & 0.1493 & 0.1124\\ 
$\tau=500$ & 0.7589 & 0.1185 & 0.0343 & 0.1450 & 0.1074\\
$\tau=700$ & 0.7986 & 0.1165 & 0.0467 & 0.1409 & \textbf{0.1033}\\
$\tau=1000$ & \textbf{0.8429} & \textbf{0.1132} & 0.0564 & \textbf{0.1349} & 0.1047\\
\hline
\end{tabular}
}
\vspace{-0.8em}
 \caption{
     \textbf{Ablation on the ID dropout start timestep $\tau$.} The time-dependent ID dropout training strategy plays a role in the tradeoff between identity preservation and pose control.}
    \label{tab:ab_tau}
    \vspace{-1.0em}
\end{table}
\begin{table}[t]
\small
\centering
\setlength\tabcolsep{2.5pt}
\resizebox{1.0\linewidth}{!}{
\begin{tabular}{l|c|cccc}
\rowcolor[gray]{.9}
\hline 
\rowcolor[gray]{.9}
Method &  $\uparrow$ ID sim. & $\downarrow$ Shape   & $\downarrow$  Pose  & $\downarrow$  Exp. & $\downarrow$ Light. \\ \hline \hline 
w/o Post-hoc Enhan. & \textbf{0.8429} & 0.1132 & 0.0564 & 0.1349 & 0.1047\\
+ w/o 3DMM model & 0.8386 & 0.1118 & 0.0427 & 0.1377 & 0.1032\\
+ w/ 3DMM model & 0.8338 & \textbf{0.1060} & \textbf{0.0358} & \textbf{0.1263} & \textbf{0.0795}\\
\hline
\end{tabular}
}
\vspace{-0.8em}
 \caption{
     \textbf{Post-hoc Head Control Enhancement at inference.} Head control metrics are boosted with the head control model.}
    \label{tab:main_infer}
\vspace{-1.3em}
\end{table}

\begin{figure}[t!]
\centering
\includegraphics[width=1.0\linewidth]{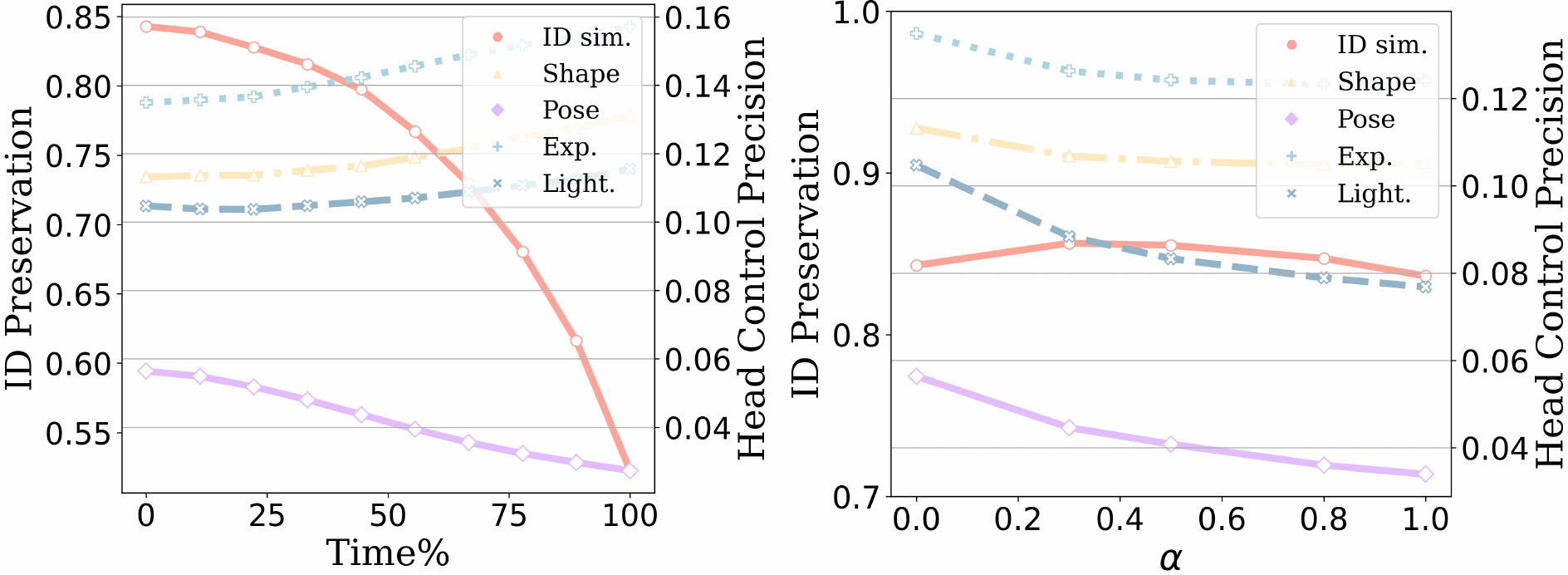}
\vspace{-1.8em}
\caption{
\textbf{Left: The utilization time~(\%) of the head control model at inference.} Using the head control model at the early stage can improve the pose control but sacrifice the identity similarity. \textbf{Right: Ablation on the control scale $\alpha$.} With the control scale $\alpha$ increasing, head control metrics are improved at a negligible cost of identity preservation.
}
\vspace{-1.8em}
\label{fig:ab_figs}
\end{figure}
\vspace{-1.4em}
\paragraph{Post-hoc Head Control Enhancement.} We further explore the possibilities of enhancing the head pose control in the inference time. We train a head control model without the identity preservation module. First, we use the head control model for the early denoising stage, and then our model with the identity preservation module. We vary the start timestep. The evaluation results are shown in Figure~\ref{fig:ab_figs}. It improves the pose metric by sacrificing the ID preservation capability. Second, we study the effect of fusion with different head control models. Specifically, we set $\pi^{\star} = \varnothing$, or w/o 3DMM model, or w/ 3DMM model in Eq.~\ref{eq:fusion}. Table~\ref{tab:main_infer} presents the results. As we can see, the pose metric further boosts when we combine our model with the head control model. Last, we perform the ablation studies on the control scale $\alpha$. Figure~\ref{fig:ab_figs} shows the head control model can strengthen the pose control at a negligible loss of identity.

% 0.8125 & 0.2056 & 0.6101 & 0.1293 & 0.0641 & 0.1519 & 0.1447 

\section{Conclusion}
In this paper, we propose a novel framework CapHuman for the human-centric image synthesis with generalizable identity preservation and fine-grained head control. We embrace the ``encode then learn to align" paradigm for generalizable identity preservation capability without further cumbersome fine-tuning. By incorporating the 3D facial representation, it enables flexible and 3D-consistent head control. Given one reference face image, our CapHuman can generate well-identity-preserved, high-fidelity, and photo-realistic human portraits with diverse head positions, poses, facial expressions, and illuminations in different contexts.

\noindent{\textbf{Acknowledgements.}} This work was supported by the National Natural Science Foundation of China (T2293723, 62293554, U2336212).

{
    \small
    \bibliographystyle{ieeenat_fullname}
    \bibliography{main}

\begin{thebibliography}{57}
\providecommand{\natexlab}[1]{#1}
\providecommand{\url}[1]{\texttt{#1}}
\expandafter\ifx\csname urlstyle\endcsname\relax
  \providecommand{\doi}[1]{doi: #1}\else
  \providecommand{\doi}{doi: \begingroup \urlstyle{rm}\Url}\fi

\bibitem[Rea(2023)]{Realistic}
Realistic vision v3.0.
\newblock \url{https://huggingface.co/SG161222/Realistic_Vision_V3.0_VAE}, 2023.

\bibitem[com(2023)]{comic-babes}
comic-babes.
\newblock \url{https://civitai.com/models/20294/comic-babes}, 2023.

\bibitem[dis(2023)]{disney-pixar-cartoon}
disney-pixar-cartoon.
\newblock \url{https://civitai.com/models/65203/disney-pixar-cartoon-type-a}, 2023.

\bibitem[too(2023)]{toonyou}
toonyou.
\newblock \url{https://civitai.com/models/30240/toonyou}, 2023.

\bibitem[Aghasanli et~al.(2023)Aghasanli, Kangin, and Angelov]{aghasanli2023interpretable}
Agil Aghasanli, Dmitry Kangin, and Plamen Angelov.
\newblock Interpretable-through-prototypes deepfake detection for diffusion models.
\newblock In \emph{ICCV}, pages 467--474, 2023.

\bibitem[Blanz and Vetter(2023)]{blanz2023morphable}
Volker Blanz and Thomas Vetter.
\newblock A morphable model for the synthesis of 3d faces.
\newblock In \emph{Seminal Graphics Papers: Pushing the Boundaries, Volume 2}, pages 157--164. 2023.

\bibitem[Blattmann et~al.(2023)Blattmann, Rombach, Ling, Dockhorn, Kim, Fidler, and Kreis]{blattmann2023align}
Andreas Blattmann, Robin Rombach, Huan Ling, Tim Dockhorn, Seung~Wook Kim, Sanja Fidler, and Karsten Kreis.
\newblock Align your latents: High-resolution video synthesis with latent diffusion models.
\newblock In \emph{CVPR}, pages 22563--22575, 2023.

\bibitem[Booth et~al.(2016)Booth, Roussos, Zafeiriou, Ponniah, and Dunaway]{booth20163d}
James Booth, Anastasios Roussos, Stefanos Zafeiriou, Allan Ponniah, and David Dunaway.
\newblock A 3d morphable model learnt from 10,000 faces.
\newblock In \emph{CVPR}, pages 5543--5552, 2016.

\bibitem[Booth et~al.(2018)Booth, Roussos, Ponniah, Dunaway, and Zafeiriou]{booth2018large}
James Booth, Anastasios Roussos, Allan Ponniah, David Dunaway, and Stefanos Zafeiriou.
\newblock Large scale 3d morphable models.
\newblock \emph{International Journal of Computer Vision}, 126\penalty0 (2):\penalty0 233--254, 2018.

\bibitem[Chang et~al.(2023)Chang, Zhang, Barber, Maschinot, Lezama, Jiang, Yang, Murphy, Freeman, Rubinstein, et~al.]{chang2023muse}
Huiwen Chang, Han Zhang, Jarred Barber, AJ Maschinot, Jose Lezama, Lu Jiang, Ming-Hsuan Yang, Kevin Murphy, William~T Freeman, Michael Rubinstein, et~al.
\newblock Muse: Text-to-image generation via masked generative transformers.
\newblock \emph{arXiv preprint arXiv:2301.00704}, 2023.

\bibitem[Corvi et~al.(2023)Corvi, Cozzolino, Zingarini, Poggi, Nagano, and Verdoliva]{Corvi_2023_ICASSP}
Riccardo Corvi, Davide Cozzolino, Giada Zingarini, Giovanni Poggi, Koki Nagano, and Luisa Verdoliva.
\newblock On the detection of synthetic images generated by diffusion models.
\newblock In \emph{ICASSP}, pages 1--5, 2023.

\bibitem[Deng et~al.(2019)Deng, Guo, Xue, and Zafeiriou]{deng2019arcface}
Jiankang Deng, Jia Guo, Niannan Xue, and Stefanos Zafeiriou.
\newblock Arcface: Additive angular margin loss for deep face recognition.
\newblock In \emph{CVPR}, pages 4690--4699, 2019.

\bibitem[Ding et~al.(2023)Ding, Zhang, Xia, Jebe, Tu, and Zhang]{ding2023diffusionrig}
Zheng Ding, Xuaner Zhang, Zhihao Xia, Lars Jebe, Zhuowen Tu, and Xiuming Zhang.
\newblock Diffusionrig: Learning personalized priors for facial appearance editing.
\newblock In \emph{CVPR}, pages 12736--12746, 2023.

\bibitem[Feng et~al.(2021)Feng, Feng, Black, and Bolkart]{DECA}
Yao Feng, Haiwen Feng, Michael~J. Black, and Timo Bolkart.
\newblock Learning an animatable detailed {3D} face model from in-the-wild images.
\newblock \emph{ACM Transactions on Graphics, (Proc. SIGGRAPH)}, 40\penalty0 (8), 2021.

\bibitem[Gal et~al.(2023)Gal, Alaluf, Atzmon, Patashnik, Bermano, Chechik, and Cohen-or]{gal2023TI}
Rinon Gal, Yuval Alaluf, Yuval Atzmon, Or Patashnik, Amit~Haim Bermano, Gal Chechik, and Daniel Cohen-or.
\newblock An image is worth one word: Personalizing text-to-image generation using textual inversion.
\newblock In \emph{ICLR}, 2023.

\bibitem[Gan et~al.(2023)Gan, Yang, Yue, Sun, and Yang]{gan2023efficient}
Yuan Gan, Zongxin Yang, Xihang Yue, Lingyun Sun, and Yi Yang.
\newblock Efficient emotional adaptation for audio-driven talking-head generation.
\newblock In \emph{CVPR}, pages 22634--22645, 2023.

\bibitem[Goodfellow et~al.(2020)Goodfellow, Pouget-Abadie, Mirza, Xu, Warde-Farley, Ozair, Courville, and Bengio]{goodfellow2020GAN}
Ian Goodfellow, Jean Pouget-Abadie, Mehdi Mirza, Bing Xu, David Warde-Farley, Sherjil Ozair, Aaron Courville, and Yoshua Bengio.
\newblock Generative adversarial networks.
\newblock \emph{Communications of the ACM}, 63\penalty0 (11):\penalty0 139--144, 2020.

\bibitem[Ho et~al.(2020)Ho, Jain, and Abbeel]{ho2020ddpm}
Jonathan Ho, Ajay Jain, and Pieter Abbeel.
\newblock Denoising diffusion probabilistic models.
\newblock \emph{NeurIPS}, 33:\penalty0 6840--6851, 2020.

\bibitem[Hu et~al.(2021)Hu, Shen, Wallis, Allen-Zhu, Li, Wang, Wang, and Chen]{hu2021lora}
Edward~J Hu, Yelong Shen, Phillip Wallis, Zeyuan Allen-Zhu, Yuanzhi Li, Shean Wang, Lu Wang, and Weizhu Chen.
\newblock Lora: Low-rank adaptation of large language models.
\newblock \emph{arXiv preprint arXiv:2106.09685}, 2021.

\bibitem[Huang et~al.(2023)Huang, Yang, Li, Yang, and Jia]{huang2023avatarfusion}
Shuo Huang, Zongxin Yang, Liangting Li, Yi Yang, and Jia Jia.
\newblock Avatarfusion: Zero-shot generation of clothing-decoupled 3d avatars using 2d diffusion.
\newblock In \emph{ACM MM}, pages 5734--5745, 2023.

\bibitem[Jia et~al.(2023)Jia, Zhao, Chan, Li, Zhang, Gong, Hou, Wang, and Su]{jia2023taming}
Xuhui Jia, Yang Zhao, Kelvin~CK Chan, Yandong Li, Han Zhang, Boqing Gong, Tingbo Hou, Huisheng Wang, and Yu-Chuan Su.
\newblock Taming encoder for zero fine-tuning image customization with text-to-image diffusion models.
\newblock \emph{arXiv preprint arXiv:2304.02642}, 2023.

\bibitem[Kang et~al.(2023)Kang, Zhu, Zhang, Park, Shechtman, Paris, and Park]{kang2023giga}
Minguk Kang, Jun-Yan Zhu, Richard Zhang, Jaesik Park, Eli Shechtman, Sylvain Paris, and Taesung Park.
\newblock Scaling up gans for text-to-image synthesis.
\newblock In \emph{CVPR}, pages 10124--10134, 2023.

\bibitem[Karras et~al.(2019)Karras, Laine, and Aila]{karras2019style}
Tero Karras, Samuli Laine, and Timo Aila.
\newblock A style-based generator architecture for generative adversarial networks.
\newblock In \emph{CVPR}, pages 4401--4410, 2019.

\bibitem[Li et~al.(2022)Li, Li, Xiong, and Hoi]{li2022blip}
Junnan Li, Dongxu Li, Caiming Xiong, and Steven Hoi.
\newblock Blip: Bootstrapping language-image pre-training for unified vision-language understanding and generation.
\newblock In \emph{ICML}, pages 12888--12900. PMLR, 2022.

\bibitem[Li et~al.(2017)Li, Bolkart, Black, Li, and Romero]{FLAME}
Tianye Li, Timo Bolkart, Michael.~J. Black, Hao Li, and Javier Romero.
\newblock Learning a model of facial shape and expression from {4D} scans.
\newblock \emph{ACM Transactions on Graphics, (Proc. SIGGRAPH Asia)}, 36\penalty0 (6):\penalty0 194:1--194:17, 2017.

\bibitem[Li et~al.(2023)Li, Ding, Zhang, Yuan, Cheng, Jiangmiao, Chen, Liu, and Loy]{li2023transformer}
Xiangtai Li, Henghui Ding, Wenwei Zhang, Haobo Yuan, Guangliang Cheng, Pang Jiangmiao, Kai Chen, Ziwei Liu, and Chen~Change Loy.
\newblock Transformer-based visual segmentation: A survey.
\newblock \emph{arXiv pre-print}, 2023.

\bibitem[Li et~al.(2024)Li, Yuan, Li, Ding, Wu, Zhang, Li, Chen, and Loy]{OMGSeg}
Xiangtai Li, Haobo Yuan, Wei Li, Henghui Ding, Size Wu, Wenwei Zhang, Yining Li, Kai Chen, and Chen~Change Loy.
\newblock Omg-seg: Is one model good enough for all segmentation?
\newblock In \emph{CVPR}, 2024.

\bibitem[Liu et~al.(2023)Liu, Wu, Van~Hoorick, Tokmakov, Zakharov, and Vondrick]{Liu_2023_zero123}
Ruoshi Liu, Rundi Wu, Basile Van~Hoorick, Pavel Tokmakov, Sergey Zakharov, and Carl Vondrick.
\newblock Zero-1-to-3: Zero-shot one image to 3d object.
\newblock In \emph{ICCV (ICCV)}, pages 9298--9309, 2023.

\bibitem[Liu et~al.(2015)Liu, Luo, Wang, and Tang]{liu2015celeba}
Ziwei Liu, Ping Luo, Xiaogang Wang, and Xiaoou Tang.
\newblock Deep learning face attributes in the wild.
\newblock In \emph{Proceedings of International Conference on Computer Vision (ICCV)}, 2015.

\bibitem[Loshchilov and Hutter(2019)]{loshchilov2017adamw}
Ilya Loshchilov and Frank Hutter.
\newblock Decoupled weight decay regularization.
\newblock In \emph{International Conference on Learning Representations}, 2019.

\bibitem[Mou et~al.(2023)Mou, Wang, Xie, Zhang, Qi, Shan, and Qie]{mou2023t2iadapter}
Chong Mou, Xintao Wang, Liangbin Xie, Jian Zhang, Zhongang Qi, Ying Shan, and Xiaohu Qie.
\newblock T2i-adapter: Learning adapters to dig out more controllable ability for text-to-image diffusion models.
\newblock \emph{arXiv preprint arXiv:2302.08453}, 2023.

\bibitem[Nichol et~al.(2021)Nichol, Dhariwal, Ramesh, Shyam, Mishkin, McGrew, Sutskever, and Chen]{nichol2021glide}
Alex Nichol, Prafulla Dhariwal, Aditya Ramesh, Pranav Shyam, Pamela Mishkin, Bob McGrew, Ilya Sutskever, and Mark Chen.
\newblock Glide: Towards photorealistic image generation and editing with text-guided diffusion models.
\newblock \emph{arXiv preprint arXiv:2112.10741}, 2021.

\bibitem[Paysan et~al.(2009)Paysan, Knothe, Amberg, Romdhani, and Vetter]{paysan2009bfm}
Pascal Paysan, Reinhard Knothe, Brian Amberg, Sami Romdhani, and Thomas Vetter.
\newblock A 3d face model for pose and illumination invariant face recognition.
\newblock In \emph{2009 sixth IEEE international conference on advanced video and signal based surveillance}, pages 296--301. Ieee, 2009.

\bibitem[Quan et~al.(2024)Quan, Wang, Tian, Ma, and Yang]{quan2024psychometry}
Ruijie Quan, Wenguan Wang, Zhibo Tian, Fan Ma, and Yi Yang.
\newblock Psychometry: An omnifit model for image reconstruction from human brain activity.
\newblock In \emph{CVPR}, 2024.

\bibitem[Radford et~al.(2021)Radford, Kim, Hallacy, Ramesh, Goh, Agarwal, Sastry, Askell, Mishkin, Clark, et~al.]{radford2021clip}
Alec Radford, Jong~Wook Kim, Chris Hallacy, Aditya Ramesh, Gabriel Goh, Sandhini Agarwal, Girish Sastry, Amanda Askell, Pamela Mishkin, Jack Clark, et~al.
\newblock Learning transferable visual models from natural language supervision.
\newblock In \emph{ICML}, pages 8748--8763. PMLR, 2021.

\bibitem[Ramesh et~al.(2021)Ramesh, Pavlov, Goh, Gray, Voss, Radford, Chen, and Sutskever]{ramesh2021dalle}
Aditya Ramesh, Mikhail Pavlov, Gabriel Goh, Scott Gray, Chelsea Voss, Alec Radford, Mark Chen, and Ilya Sutskever.
\newblock Zero-shot text-to-image generation.
\newblock In \emph{ICML}, pages 8821--8831. PMLR, 2021.

\bibitem[Ramesh et~al.(2022)Ramesh, Dhariwal, Nichol, Chu, and Chen]{ramesh2022dalle2}
Aditya Ramesh, Prafulla Dhariwal, Alex Nichol, Casey Chu, and Mark Chen.
\newblock Hierarchical text-conditional image generation with clip latents.
\newblock \emph{arXiv preprint arXiv:2204.06125}, 1\penalty0 (2):\penalty0 3, 2022.

\bibitem[Rombach et~al.(2022)Rombach, Blattmann, Lorenz, Esser, and Ommer]{rombach2022SD}
Robin Rombach, Andreas Blattmann, Dominik Lorenz, Patrick Esser, and Bj{\"o}rn Ommer.
\newblock High-resolution image synthesis with latent diffusion models.
\newblock In \emph{CVPR}, pages 10684--10695, 2022.

\bibitem[Ruiz et~al.(2023)Ruiz, Li, Jampani, Pritch, Rubinstein, and Aberman]{ruiz2023dreambooth}
Nataniel Ruiz, Yuanzhen Li, Varun Jampani, Yael Pritch, Michael Rubinstein, and Kfir Aberman.
\newblock Dreambooth: Fine tuning text-to-image diffusion models for subject-driven generation.
\newblock In \emph{CVPR}, pages 22500--22510, 2023.

\bibitem[Saharia et~al.(2022)Saharia, Chan, Saxena, Li, Whang, Denton, Ghasemipour, Gontijo~Lopes, Karagol~Ayan, Salimans, et~al.]{saharia2022imagen}
Chitwan Saharia, William Chan, Saurabh Saxena, Lala Li, Jay Whang, Emily~L Denton, Kamyar Ghasemipour, Raphael Gontijo~Lopes, Burcu Karagol~Ayan, Tim Salimans, et~al.
\newblock Photorealistic text-to-image diffusion models with deep language understanding.
\newblock \emph{NeurIPS}, 35:\penalty0 36479--36494, 2022.

\bibitem[Schroff et~al.(2015)Schroff, Kalenichenko, and Philbin]{schroff2015facenet}
Florian Schroff, Dmitry Kalenichenko, and James Philbin.
\newblock Facenet: A unified embedding for face recognition and clustering.
\newblock In \emph{CVPR}, pages 815--823, 2015.

\bibitem[Schuhmann et~al.(2022)Schuhmann, Beaumont, Vencu, Gordon, Wightman, Cherti, Coombes, Katta, Mullis, Wortsman, et~al.]{schuhmann2022laion}
Christoph Schuhmann, Romain Beaumont, Richard Vencu, Cade Gordon, Ross Wightman, Mehdi Cherti, Theo Coombes, Aarush Katta, Clayton Mullis, Mitchell Wortsman, et~al.
\newblock Laion-5b: An open large-scale dataset for training next generation image-text models.
\newblock \emph{NeurIPS}, 35:\penalty0 25278--25294, 2022.

\bibitem[Shen et~al.(2024)Shen, Ma, Zhou, and Yang]{shen2023controllable}
Xiaolong Shen, Jianxin Ma, Chang Zhou, and Zongxin Yang.
\newblock Controllable 3d face generation with conditional style code diffusion.
\newblock In \emph{AAAI}, 2024.

\bibitem[Shi et~al.(2023)Shi, Xiong, Lin, and Jung]{shi2023instantbooth}
Jing Shi, Wei Xiong, Zhe Lin, and Hyun~Joon Jung.
\newblock Instantbooth: Personalized text-to-image generation without test-time finetuning.
\newblock \emph{arXiv preprint arXiv:2304.03411}, 2023.

\bibitem[Vaswani et~al.(2017)Vaswani, Shazeer, Parmar, Uszkoreit, Jones, Gomez, Kaiser, and Polosukhin]{vaswani2017attention}
Ashish Vaswani, Noam Shazeer, Niki Parmar, Jakob Uszkoreit, Llion Jones, Aidan~N Gomez, {\L}ukasz Kaiser, and Illia Polosukhin.
\newblock Attention is all you need.
\newblock \emph{NeurIPS}, 30, 2017.

\bibitem[Xiao et~al.(2023)Xiao, Yin, Freeman, Durand, and Han]{xiao2023fastcomposer}
Guangxuan Xiao, Tianwei Yin, William~T. Freeman, Frédo Durand, and Song Han.
\newblock Fastcomposer: Tuning-free multi-subject image generation with localized attention.
\newblock \emph{arXiv}, 2023.

\bibitem[Xu et~al.(2023{\natexlab{a}})Xu, Wang, Cheng, Cao, Shan, Qie, and Gao]{xu2023dream3d}
Jiale Xu, Xintao Wang, Weihao Cheng, Yan-Pei Cao, Ying Shan, Xiaohu Qie, and Shenghua Gao.
\newblock Dream3d: Zero-shot text-to-3d synthesis using 3d shape prior and text-to-image diffusion models.
\newblock In \emph{CVPR}, pages 20908--20918, 2023{\natexlab{a}}.

\bibitem[Xu et~al.(2023{\natexlab{b}})Xu, Yang, and Yang]{xu2023seeavatar}
Yuanyou Xu, Zongxin Yang, and Yi Yang.
\newblock Seeavatar: Photorealistic text-to-3d avatar generation with constrained geometry and appearance.
\newblock \emph{arXiv preprint arXiv:2312.08889}, 2023{\natexlab{b}}.

\bibitem[Yang et~al.(2021)Yang, Zhuang, and Pan]{yang2021multiple}
Yi Yang, Yueting Zhuang, and Yunhe Pan.
\newblock Multiple knowledge representation for big data artificial intelligence: framework, applications, and case studies.
\newblock \emph{Frontiers of Information Technology \& Electronic Engineering}, 22\penalty0 (12):\penalty0 1551--1558, 2021.

\bibitem[Yang et~al.(2024)Yang, Chen, Li, Wang, and Yang]{yang2024doraemongpt}
Zongxin Yang, Guikun Chen, Xiaodi Li, Wenguan Wang, and Yi Yang.
\newblock Doraemongpt: Toward understanding dynamic scenes with large language models.
\newblock \emph{arXiv preprint arXiv:2401.08392}, 2024.

\bibitem[Ye et~al.(2023)Ye, Zhang, Liu, Han, and Yang]{ye2023ip-adapter}
Hu Ye, Jun Zhang, Sibo Liu, Xiao Han, and Wei Yang.
\newblock Ip-adapter: Text compatible image prompt adapter for text-to-image diffusion models.
\newblock 2023.

\bibitem[Yu et~al.(2022)Yu, Xu, Koh, Luong, Baid, Wang, Vasudevan, Ku, Yang, Ayan, et~al.]{yu2022parti}
Jiahui Yu, Yuanzhong Xu, Jing~Yu Koh, Thang Luong, Gunjan Baid, Zirui Wang, Vijay Vasudevan, Alexander Ku, Yinfei Yang, Burcu~Karagol Ayan, et~al.
\newblock Scaling autoregressive models for content-rich text-to-image generation.
\newblock \emph{arXiv preprint arXiv:2206.10789}, 2\penalty0 (3):\penalty0 5, 2022.

\bibitem[Zhang et~al.(2023)Zhang, Rao, and Agrawala]{zhang2023controlnet}
Lvmin Zhang, Anyi Rao, and Maneesh Agrawala.
\newblock Adding conditional control to text-to-image diffusion models.
\newblock In \emph{ICCV}, 2023.

\bibitem[Zhang et~al.(2024)Zhang, Yang, and Yang]{zhang2023sifu}
Zechuan Zhang, Zongxin Yang, and Yi Yang.
\newblock Sifu: Side-view conditioned implicit function for real-world usable clothed human reconstruction.
\newblock In \emph{CVPR}, 2024.

\bibitem[Zhou et~al.(2023)Zhou, Yang, and Yang]{zhou2023pyramid}
Dewei Zhou, Zongxin Yang, and Yi Yang.
\newblock Pyramid diffusion models for low-light image enhancement.
\newblock In \emph{IJCAI}, 2023.

\bibitem[Zhou et~al.(2024{\natexlab{a}})Zhou, Li, Ma, Yang, and Yang]{zhou2024migc}
Dewei Zhou, You Li, Fan Ma, Zongxin Yang, and Yi Yang.
\newblock Migc: Multi-instance generation controller for text-to-image synthesis.
\newblock In \emph{CVPR}, 2024{\natexlab{a}}.

\bibitem[Zhou et~al.(2024{\natexlab{b}})Zhou, Ma, Fan, and Yang]{zhou2024headstudio}
Zhenglin Zhou, Fan Ma, Hehe Fan, and Yi Yang.
\newblock Headstudio: Text to animatable head avatars with 3d gaussian splatting.
\newblock \emph{arXiv preprint arXiv:2402.06149}, 2024{\natexlab{b}}.

\end{thebibliography}
}

\clearpage
\setcounter{page}{1}
\maketitlesupplementary

\section{More Qualitative Results}
\begin{figure}[t!]
\centering
\includegraphics[width=1.0\linewidth]{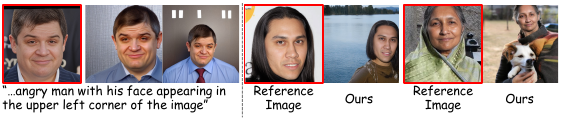}
\vspace{-1.9em}
\caption{
\textbf{Left:} The detailed prompt struggles to control the head position and facial expression. \textbf{Right:} Maintain the hairstyle.
}
\vspace{-0.6em}
\label{fig:detail_prompt_hairstyle}
\end{figure}
\paragraph{Can the detailed prompt achieve the head control as well?} In Figure~\ref{fig:detail_prompt_hairstyle}~(Left), we show the detailed prompt still struggles to control the human head, \eg position, and facial expression.

\paragraph{Maintain the hairstyle.} In Figure~\ref{fig:detail_prompt_hairstyle}~(Right), we show that our model can keep the hairstyle via minor modifications, that is, keep the hair area in the ID features and masks.

\begin{figure}[t!]
\centering
\includegraphics[width=1.0\linewidth]{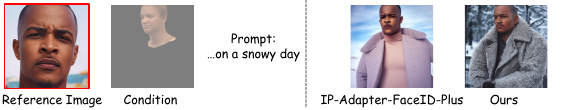}
\vspace{-1.9em}
\caption{
\textbf{Visual comparison with IP-Adapter.} Our method shows better ID preservation and head control while following the given prompt.
}
\vspace{-0.6em}
\label{fig:vis_ip_adapter}
\end{figure}

\paragraph{Visual comparison with IP-Adapter.} As shown in Figure~\ref{fig:vis_ip_adapter}, we compare our method with IP-Adapter~\cite{ye2023ip-adapter}. Our method shows better ID preservation and head control while following the given prompt.

\begin{figure*}[t!]
\centering
\includegraphics[width=1.0\textwidth]{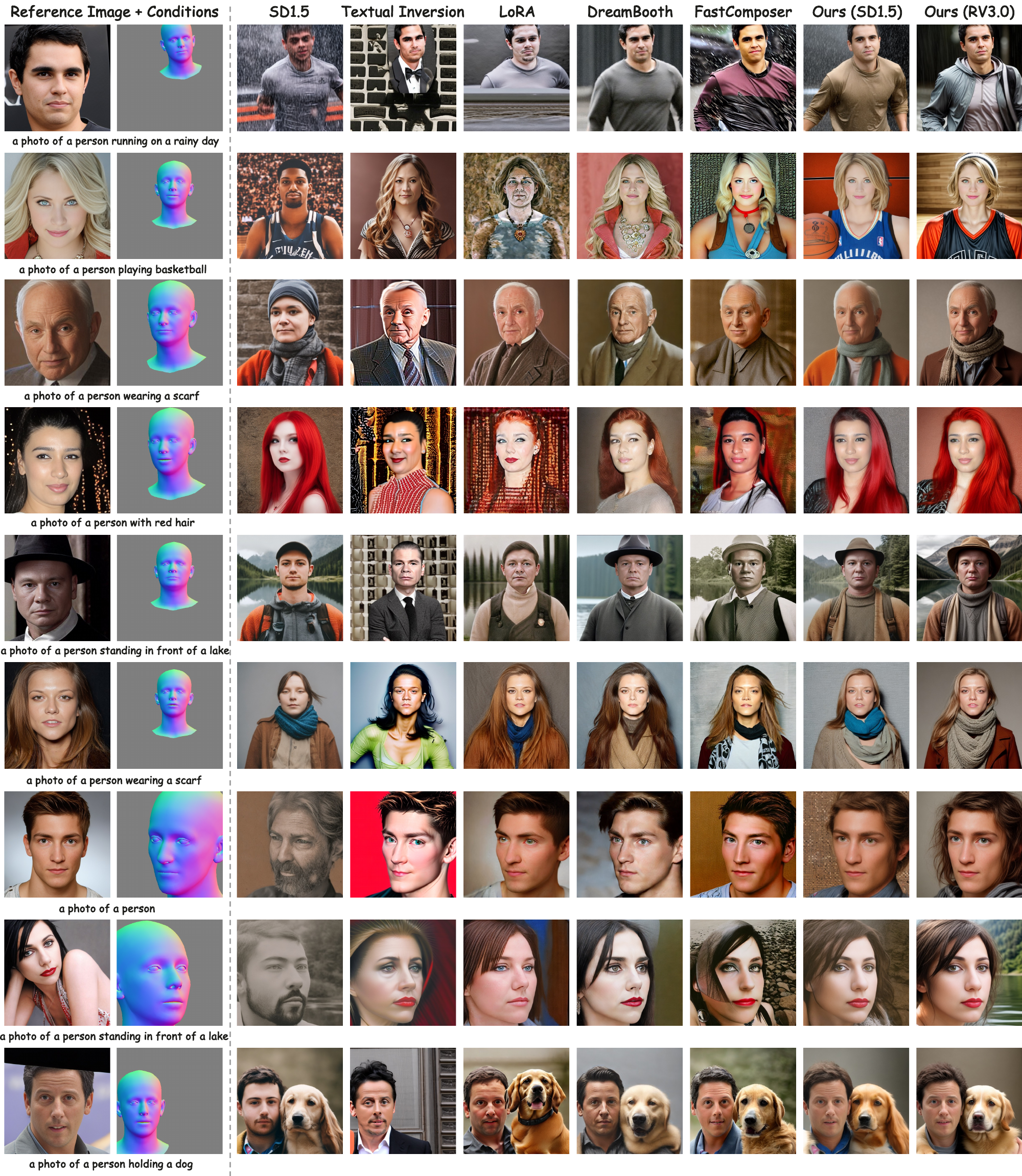}
\vspace{-1.9em}
\caption{
\textbf{More qualitative results.} Our CapHuman can produce well-identity-preserved, photo-realistic, and high-fidelity portraits with various head positions and poses in different contexts, compared with the baselines. 
Note that our model can be combined with other pre-trained models, \eg RealisticVision~\cite{Realistic} in the community flexibly. For the head condition, we only display the Surface Normal here.
}
\vspace{-1.5em}
\label{fig:supp_main_results1}
\end{figure*}

%%%%%%%%%%%%%%%%%%%%%%%%%%
% if size > 20 MB, remove this figure~\ref{fig:supp_main_results2}
%%%%%%%%%%%%%%%%%%%%%%%%%%
% \paragraph{Visual comparisons.} We show more visual comparisons with the established baselines~\cite{rombach2022SD, gal2023TI, ruiz2023dreambooth, hu2021lora, xiao2023fastcomposer} in Figure~\ref{fig:supp_main_results1} and~\ref{fig:supp_main_results2}. Our CapHuman can generate well-identity-preserved, photo-realistic, and high-fidelity portraits with various head positions and poses in different contexts.
\paragraph{Visual comparisons.} We show more visual comparisons with the established baselines~\cite{rombach2022SD, gal2023TI, ruiz2023dreambooth, hu2021lora, xiao2023fastcomposer} in Figure~\ref{fig:supp_main_results1}. Our CapHuman can generate well-identity-preserved, photo-realistic, and high-fidelity portraits with various head positions and poses in different contexts.

\begin{figure*}[t!]
\centering
\includegraphics[width=1.0\textwidth]{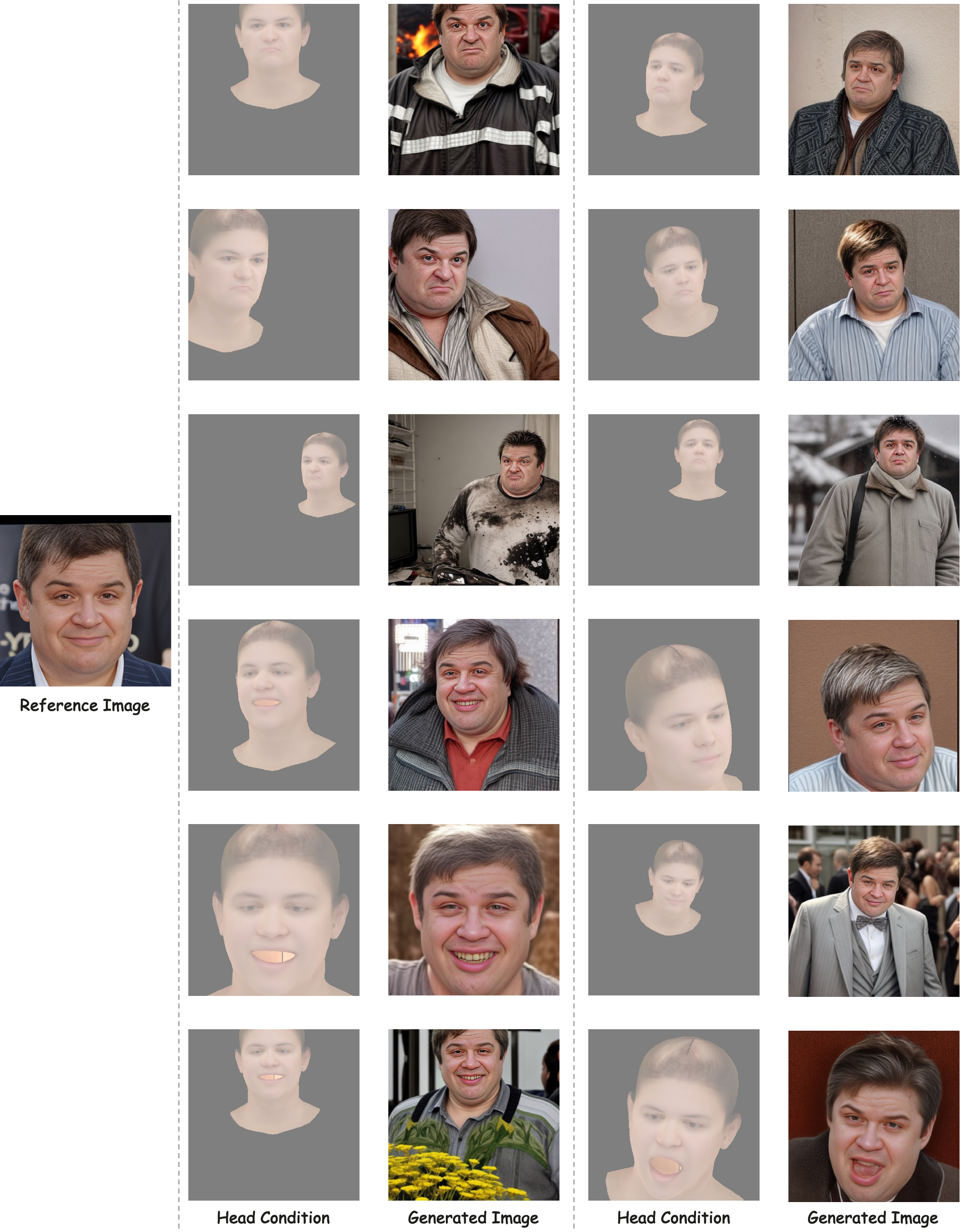}
\vspace{-1.9em}
\caption{
\textbf{More results with different and rich facial expressions.} Our CapHuman can provide facial expression control in a flexible and fine-grained manner.
}
\label{fig:supp_exp}
\end{figure*}

\paragraph{Facial expression control.} In Figure~\ref{fig:supp_exp}, we provide more examples, demonstrating the facial expression control ability of our CapHuman.

\section{More Quantitative Results}
\begin{table}[t]
\small
\centering
\setlength\tabcolsep{6.5pt}
\resizebox{1.0\linewidth}{!}{
\begin{tabular}{l|c|c|c}
\rowcolor[gray]{.9}
\hline 
Method &  \#ref. & $\uparrow$ ID sim. & $\downarrow$ Personalization time~(s)  \\ \hline \hline 
LoRA~\cite{hu2021lora} & 5 & 0.6298 & 1223\\ 
DreamBooth~\cite{ruiz2023dreambooth} & 5 & 0.7457 & 1321 \\
Ours & 1 & \textbf{0.8429} & \textbf{7} \\
\hline
\end{tabular}
}
\vspace{-0.8em}
 \caption{
     \textbf{Comparisons with fine-tuning methods with more reference images.} Ours still outperforms other baselines with higher identity similarity and faster speed.}
    \label{tab:ab_ref}
\vspace{-1.5em}
\end{table}
\paragraph{More reference images.} We compare our method with fine-tuning methods that take more reference images as input. The results are presented in Table~\ref{tab:ab_ref}. Our method still outperforms LoRA~\cite{hu2021lora} and DreamBooth~\cite{ruiz2023dreambooth} with better identity preservation and shorter personalization time.
\begin{figure}[t!]
\centering
\includegraphics[width=1.0\linewidth]{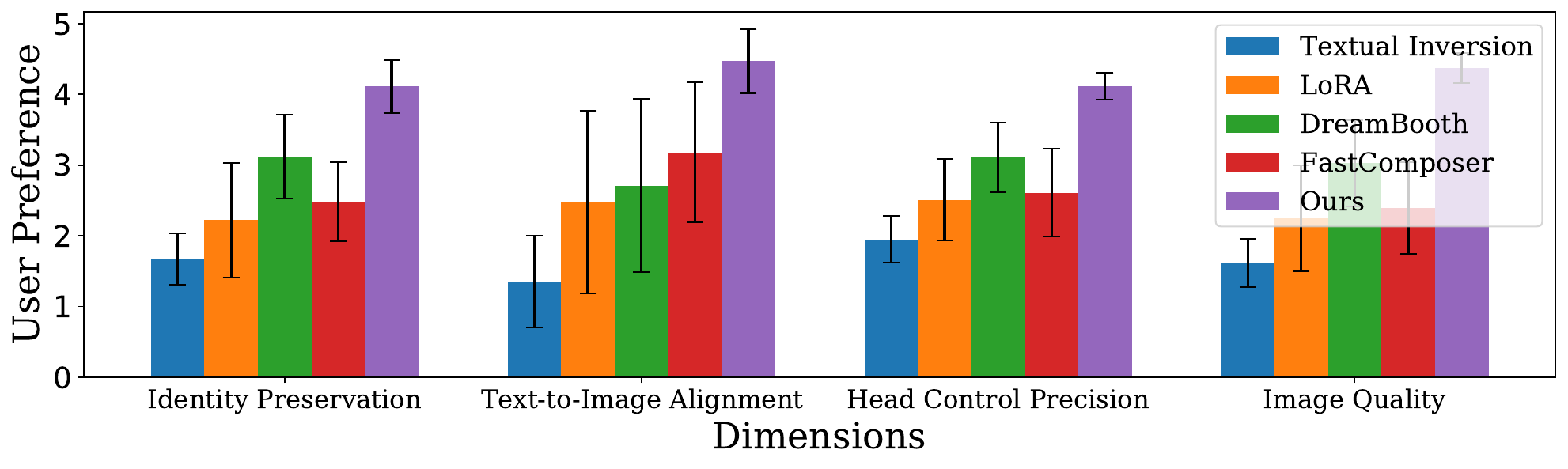}
\vspace{-2.0em}
\caption{
\textbf{User Study.} Users prefer our method in all four dimensions: identity preservation, text-to-image alignment, head control precision, and image quality.
}
\vspace{-1.0em}
\label{fig:ab_userstudy}
\end{figure}
\paragraph{User Study.} We invite 50 users to score 20 groups of results from each method in terms of the following four dimensions: identity preservation, text-to-image alignment, head control precision, and image quality. Figure~\ref{fig:ab_userstudy} shows our method is much more preferred by the users.
\begin{table}[t]
\centering
\setlength\tabcolsep{3pt}
\resizebox{1.0\linewidth}{!}{
\begin{tabular}{l|c|cc|cccc}
\hline
\rowcolor[gray]{0.9}
Method &  $\uparrow$ ID sim.   & $\uparrow$  CLIP score  & $\uparrow$  Prompt acc. & $\downarrow$ Shape   & $\downarrow$  Pose  & $\downarrow$  Exp.& $\downarrow$ Light. \\ \hline \hline 
IP-Adapter-FaceID-Plus~\cite{ye2023ip-adapter} & 0.8125 & 0.2056 & 61.01\% & 0.1293 & 0.0641 & 0.1519 & 0.1447 \\
Ours & \textbf{0.8363} & \textbf{0.2256} & \textbf{74.17\%} & \textbf{0.1020} & \textbf{0.0436} & \textbf{0.1241} & \textbf{0.0965}\\
\hline
\end{tabular}
}
\vspace{-0.8em}
 \caption{\textbf{Comparison with IP-Adapter.} Our method outperforms IP-Adapter in all aspects. }
    \label{tab:comp_ip_adapter}
\vspace{-1.3em}
\end{table}

\paragraph{Comparison with IP-Adapter.} We compare our method with IP-Adapter~\cite{ye2023ip-adapter}. The results are presented in Table~\ref{tab:comp_ip_adapter}. Our method outperforms IP-Adapter~\cite{ye2023ip-adapter} in all aspects.
\paragraph{Ablation on the global ID feature.} For the choice of the global ID feature, we compare FaceNet~\cite{schroff2015facenet} and ArcFace~\cite{deng2019arcface}. \textcolor[RGB]{0,176,80}{FaceNet} outperforms \textcolor[RGB]{192,0,0}{ArcFace}. The ID similarity is \textcolor[RGB]{0,176,80}{0.8367} (\textcolor[RGB]{192,0,0}{0.8091}) measured by FaceNet and \textcolor[RGB]{0,176,80}{0.4819} (\textcolor[RGB]{192,0,0}{0.4737}) measured by ArcFace.

\section{More Applications}
\begin{figure*}[t!]
\centering
\includegraphics[width=1.0\textwidth]{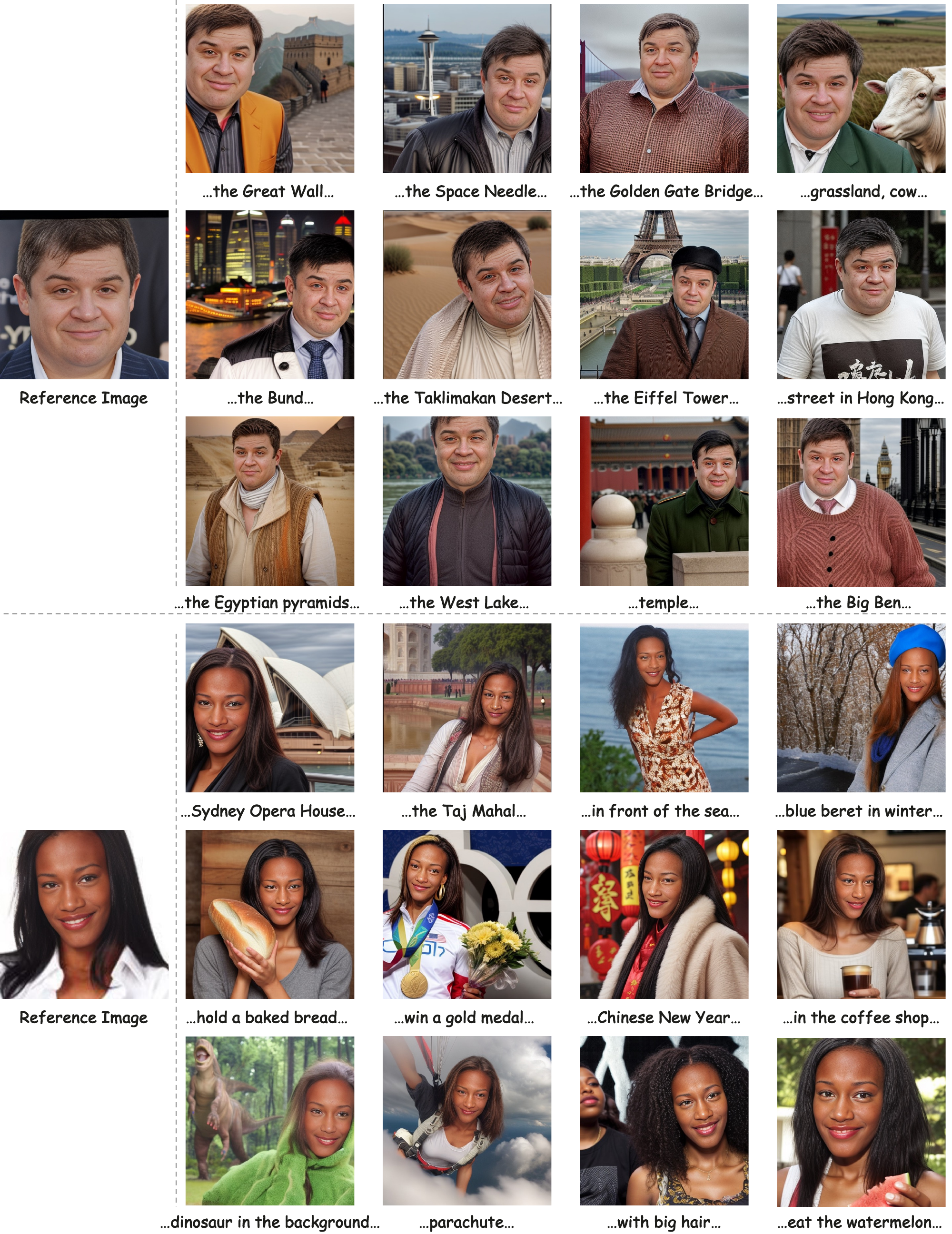}
\vspace{-1.9em}
\caption{
\textbf{More results in the realistic style.} Our CapHuman can be adapted to produce various identity-preserved and photo-realistic portraits with diverse head positions, poses, and facial expressions.
}
\label{fig:supp_real}
\end{figure*}

\begin{figure*}[t!]
\centering
\includegraphics[width=1.0\textwidth]{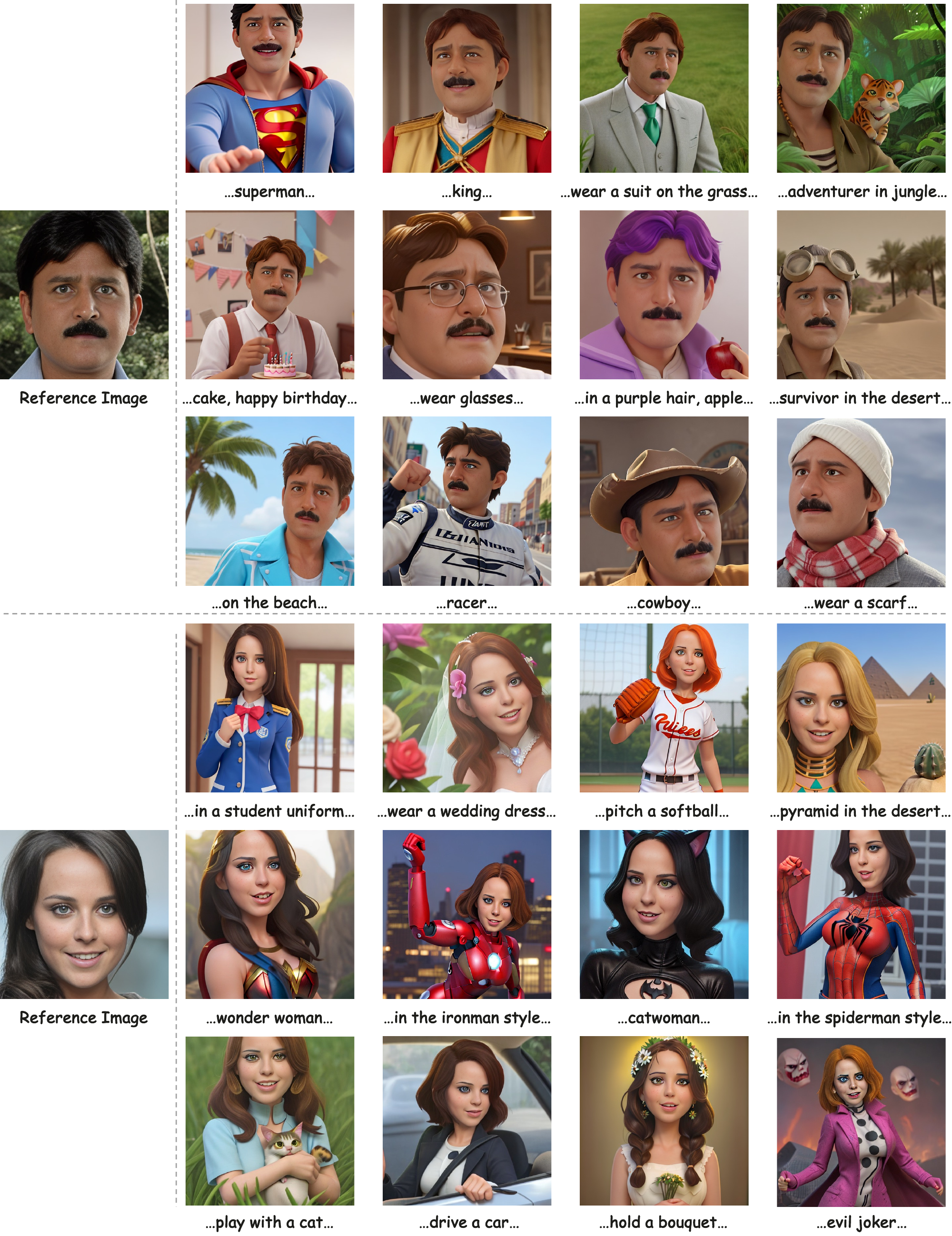}
\vspace{-1.9em}
\caption{
\textbf{More results in the Disney cartoon style.} Our CapHuman can be adapted to produce various identity-preserved portraits with diverse head positions, poses, and facial expressions.
}
\label{fig:supp_disney}
\end{figure*}

\begin{figure*}[t!]
\centering
\includegraphics[width=1.0\textwidth]{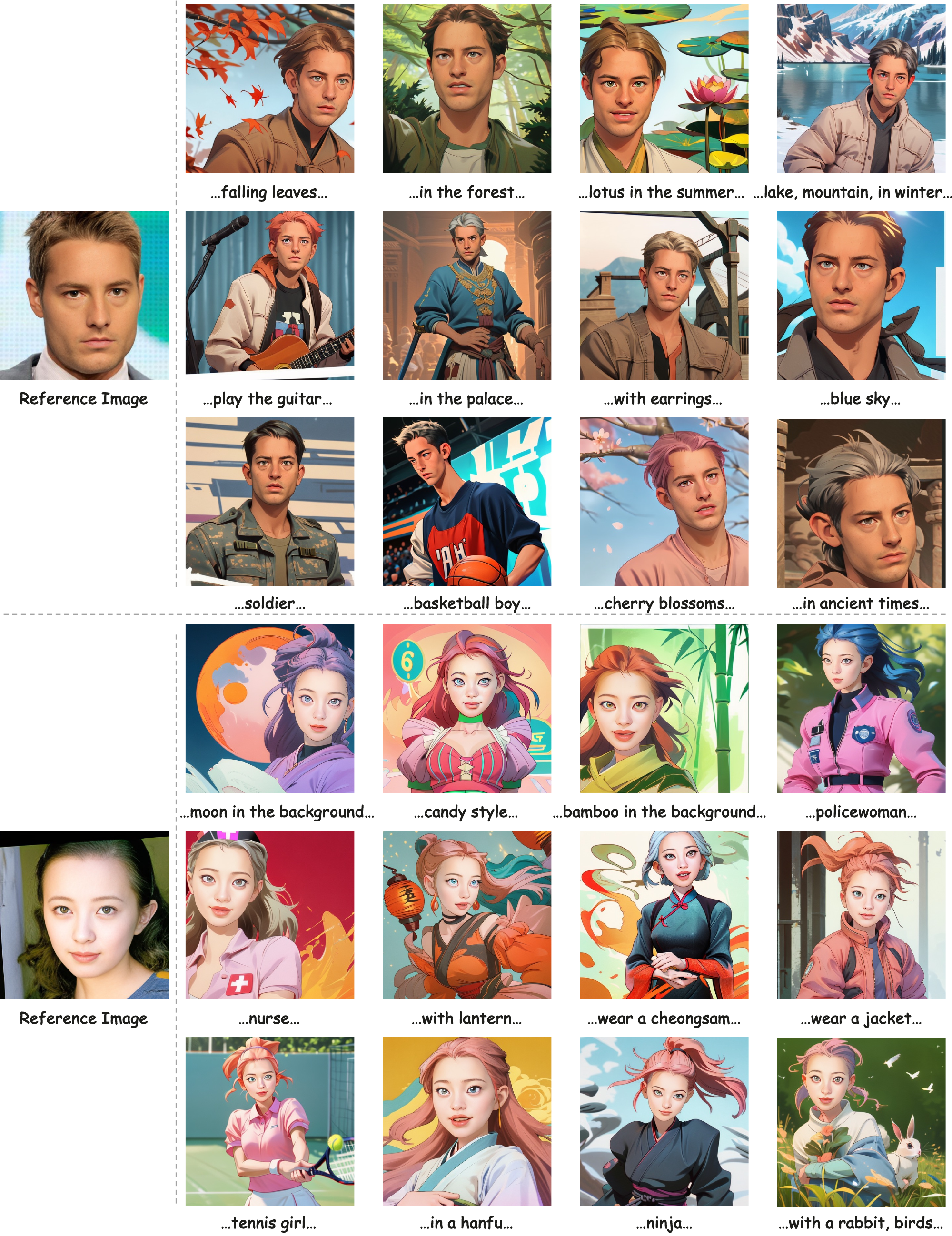}
\vspace{-1.9em}
\caption{
\textbf{More results in the animation style.} Our CapHuman can be adapted to produce various identity-preserved portraits with diverse head positions, poses, and facial expressions.
}
\label{fig:supp_toon}
\end{figure*}

\begin{figure*}[t!]
\centering
\includegraphics[width=1.0\textwidth]{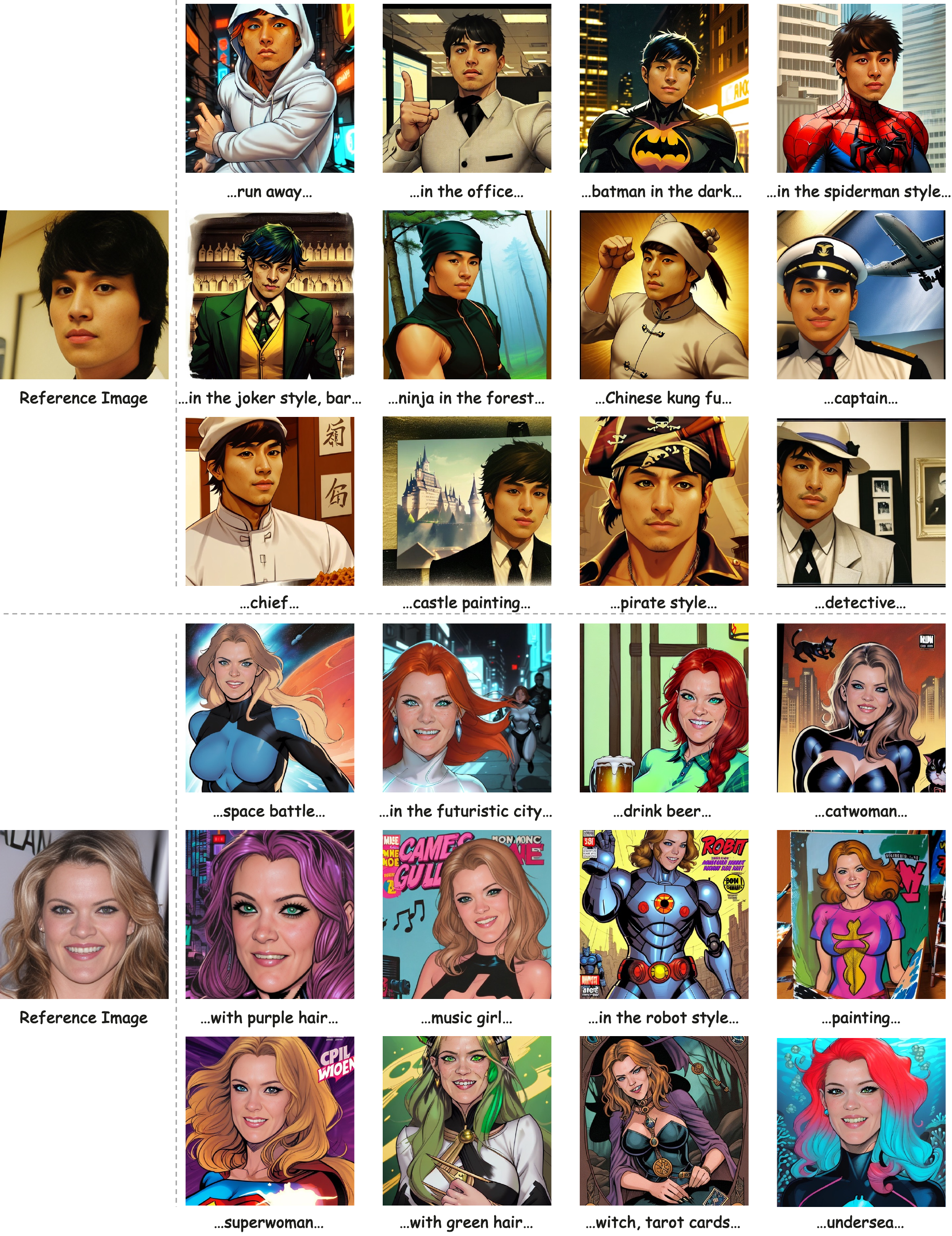}
\vspace{-1.9em}
\caption{
\textbf{More results in the comic style.} Our CapHuman can be adapted to produce various identity-preserved portraits with diverse head positions, poses, and facial expressions.
}
\label{fig:supp_comic}
\end{figure*}

\paragraph{Stylization by adaptation to other pre-trained models.} Benefitting from the nature of open-source in the community, we can inherit the rich pre-trained models. Our CapHuman can be adapted to other pre-trained models~\cite{Realistic, toonyou, disney-pixar-cartoon, comic-babes} in the community flexibly, which can generate identity-preserved portraits with various head positions, poses, and facial expressions in different styles. More results are presented in Figure~\ref{fig:supp_real},~\ref{fig:supp_disney},~\ref{fig:supp_toon}, and~\ref{fig:supp_comic}.

\begin{figure*}[t!]
\centering
\includegraphics[width=1.0\textwidth]{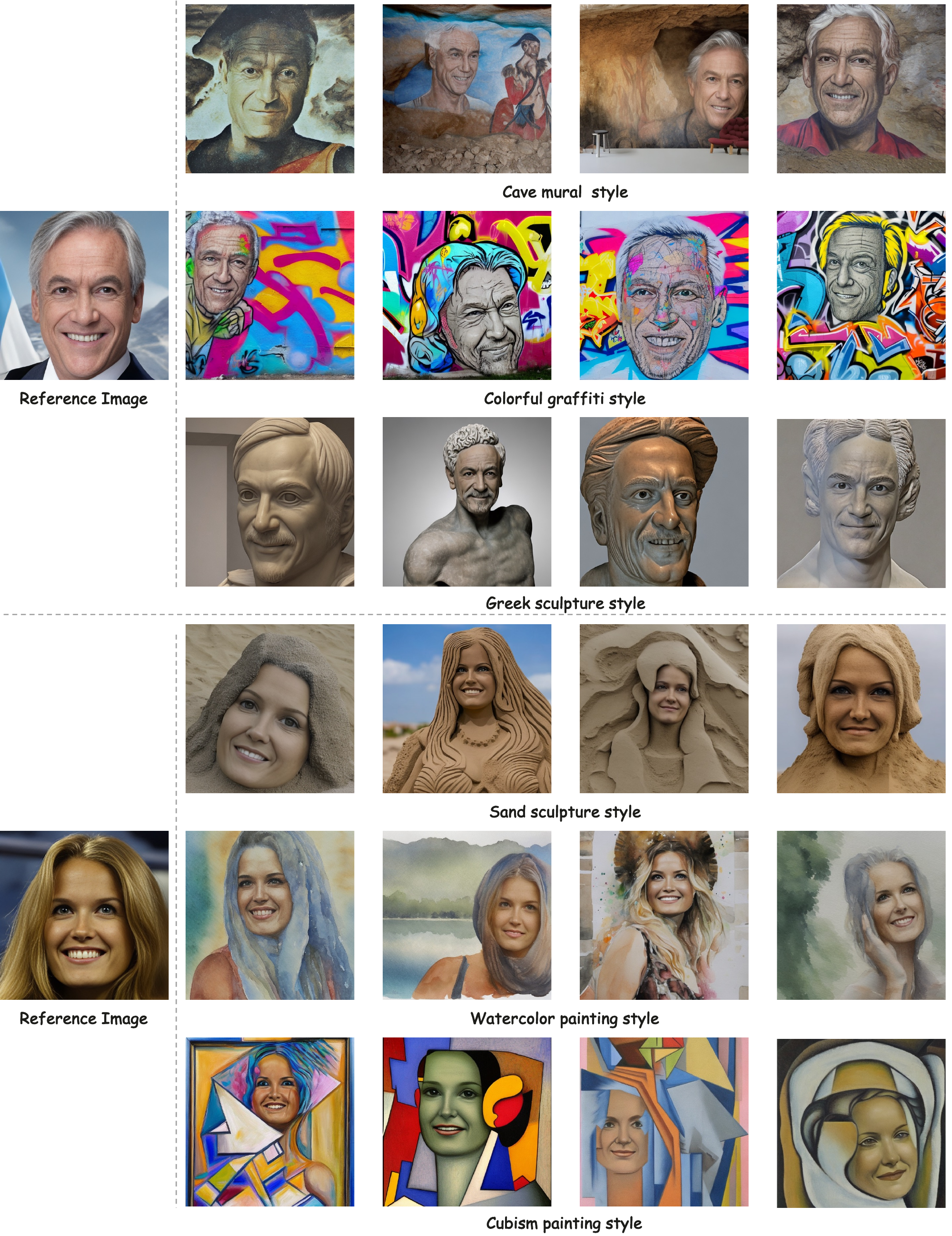}
\vspace{-1.9em}
\caption{
\textbf{Stylization by style prompts.} Our CapHuman can generate identity-preserved portraits with different styles by style prompts. 
}
\label{fig:supp_prompt}
\end{figure*}

\paragraph{Stylization by style prompts.} We also showcase portraits with different styles driven by style prompts in Figure~\ref{fig:supp_prompt}.

\begin{figure*}[t!]
\centering
\includegraphics[width=1.0\textwidth]{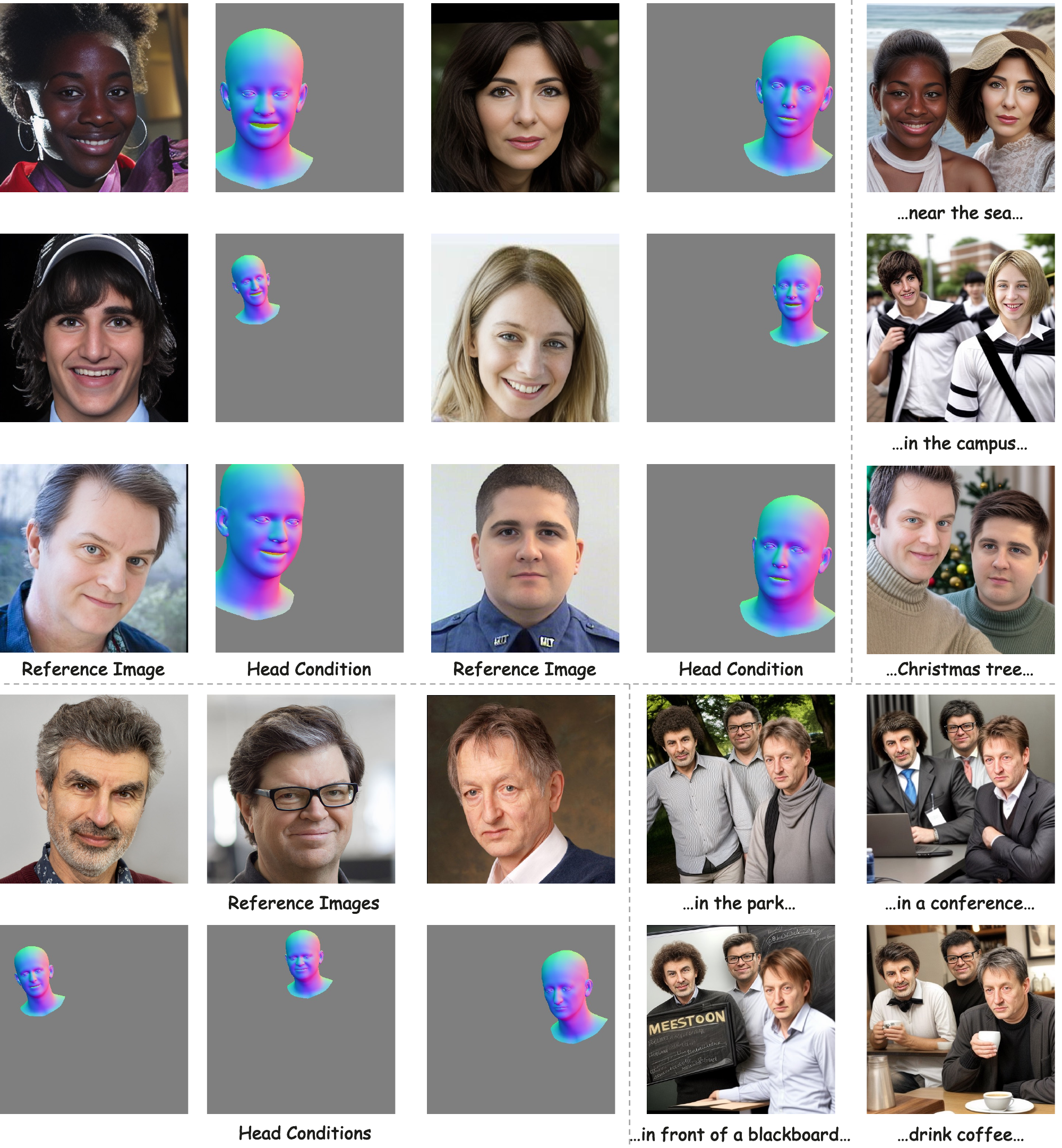}
\vspace{-1.9em}
\caption{
\textbf{Multi-Human image generation.} Given reference images, our CapHuman can generate various identity-preserved multi-human images, consistent with the corresponding head conditions.
}
\label{fig:supp_multiperson}
\end{figure*}

\paragraph{Multi-Human image generation.} Our CapHuman supports multi-human image generation. The generated results are presented in Figure~\ref{fig:supp_multiperson}.

\begin{figure*}[t!]
\centering
\includegraphics[width=1.0\textwidth]{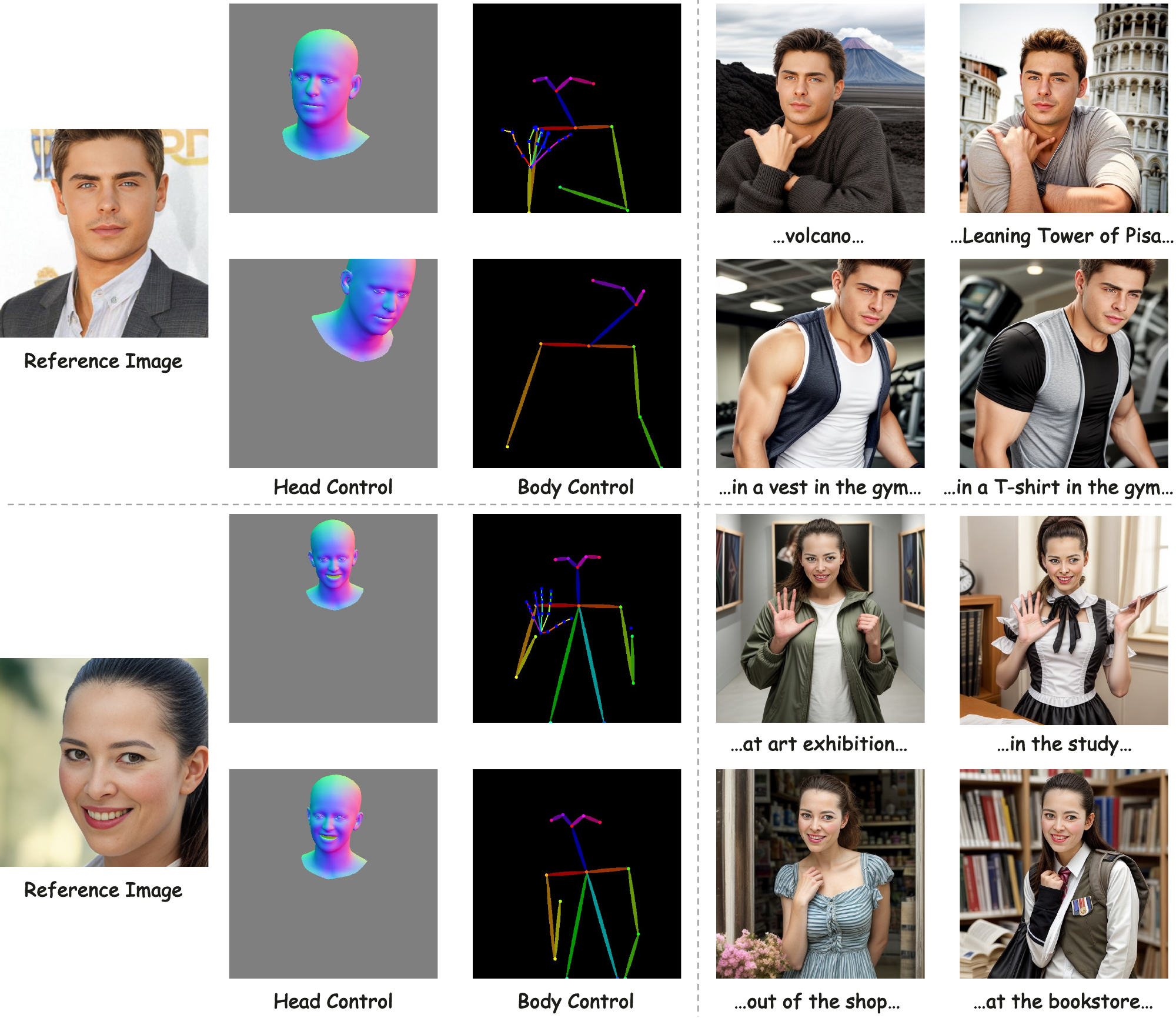}
\vspace{-1.9em}
\caption{
\textbf{Simultaneous head and body control with identity preservation.} Our CapHuman can control the head and body simultaneously with the pose-guided ControlNet~\cite{zhang2023controlnet} with identity preservation.
}
\label{fig:supp_controlnet}
\end{figure*}

\paragraph{Simultaneous head and body control.} Combined with the pose-guided ControlNet~\cite{zhang2023controlnet}, our CapHuman can control the head and the body simultaneously with identity preservation. More results are presented in Figure~\ref{fig:supp_controlnet}.

\begin{figure*}[t!]
\centering
\includegraphics[width=1.0\textwidth]{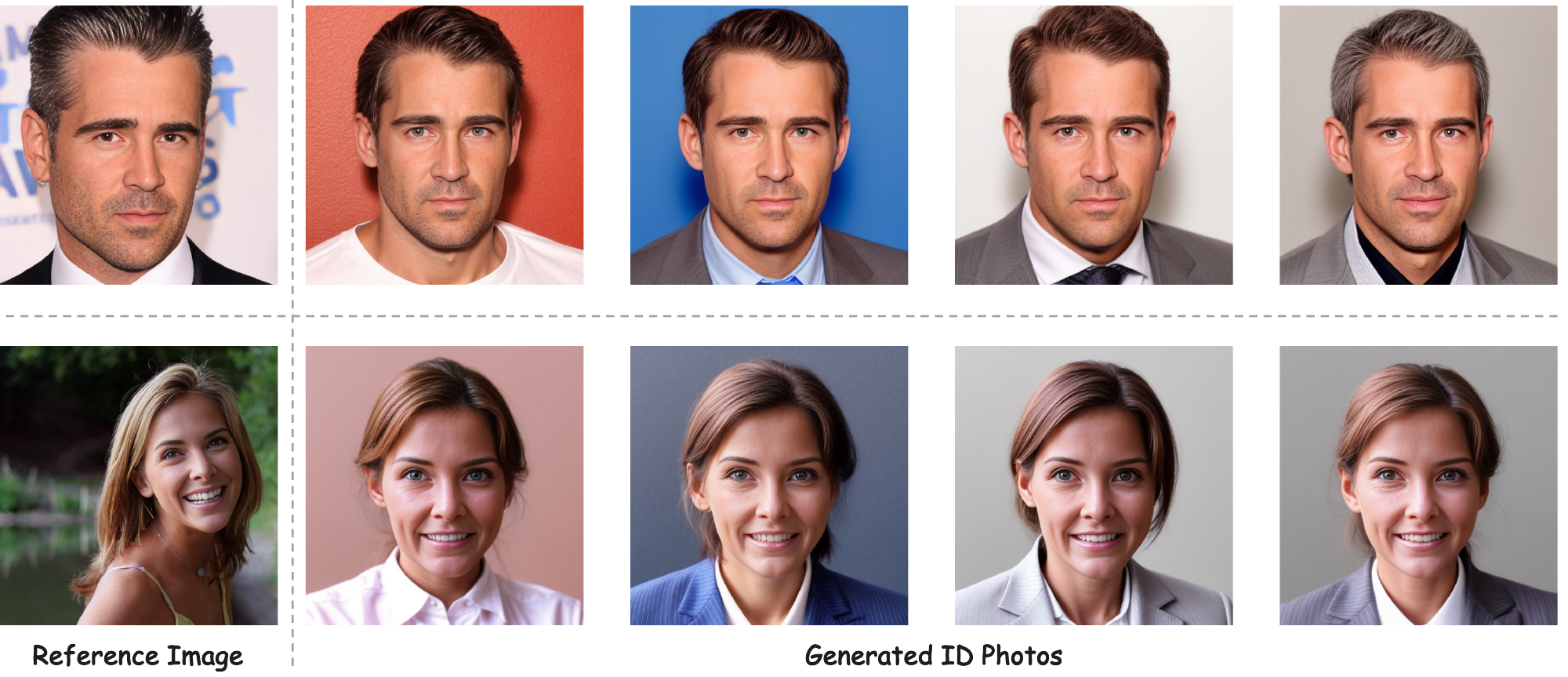}
\vspace{-1.9em}
\caption{
\textbf{Photo ID generation.} Our CapHuman can generate standard ID photos by adjusting the head conditions and providing the proper prompts.
}
\label{fig:supp_idphoto}
\end{figure*}

\paragraph{Photo ID generation.} Photo ID is widely used in passports, ID cards, etc. There are typically some requirements for these photos, \eg plain background, formal wearing, and standard head pose. As shown in Figure~\ref{fig:supp_idphoto}, our CapHuman can generate standard ID photos by adjusting the head conditions and providing the proper prompts conveniently.

\section{HumanIPHC Benchmark Details}
We introduce more details about our HumanIPHC benchmark in this section.
\paragraph{ID split.} 100 IDs used in our benchmark are listed in Table~\ref{tab:id_list}.

\paragraph{Prompts.} We list the prompts used in the benchmark:
\begin{itemize}
    \item a photo of a person.
    \item a photo of a person with red hair.
    \item a photo of a person standing in front of a lake.
    \item a photo of a person holding a dog.
    \item a photo of a person running on a rainy day.
    \item a closeup of a person playing the guitar.
    \item a photo of a person wearing a suit on a snowy day.
    \item a photo of a person playing basketball.
    \item a photo of a person wearing a scarf.
    \item a photo of a person on a cobblestone street.
    \item a photo of a person with a sheep in the background.
    \item a photo of a person sitting on a purple rug in a forest.
    \item a photo of a person with a tree and autumn leaves in the background.
    \item a photo of a person with the Eiffel Tower in the background.
    \item a photo of a person wearing a red sweater.
    \item a photo of a person wearing a spacesuit.
    \item a photo of a person wearing a green coat.
    \item a photo of a person wearing a blue hoodie.
    \item a photo of a person wearing a santa hat.
    \item a photo of a person wearing a yellow shirt.
    \item a photo of a person with a city in the background.
    \item a photo of a person with a mountain in the background.
    \item a photo of a person on the beach.
    \item a photo of a person in the jungle.
    \item a photo of a person riding a horse.
    \item a photo of a person holding a bottle of a red wine.
    \item a photo of a person swimming in the pool.
    \item a photo of a person holding flowers.
    \item a photo of a person with a cat.
    \item a photo of a person reading a book.
    \item a photo of a person in a chief outfit.
    \item a photo of a person in a police outfit.
    \item a photo of a person in a firefighter outfit.
    \item a photo of a person in a purple wizard outfit.
    \item a photo of a person wearing a necklace.
\end{itemize}

\begin{figure*}[t!]
\centering
\includegraphics[width=1.0\textwidth]{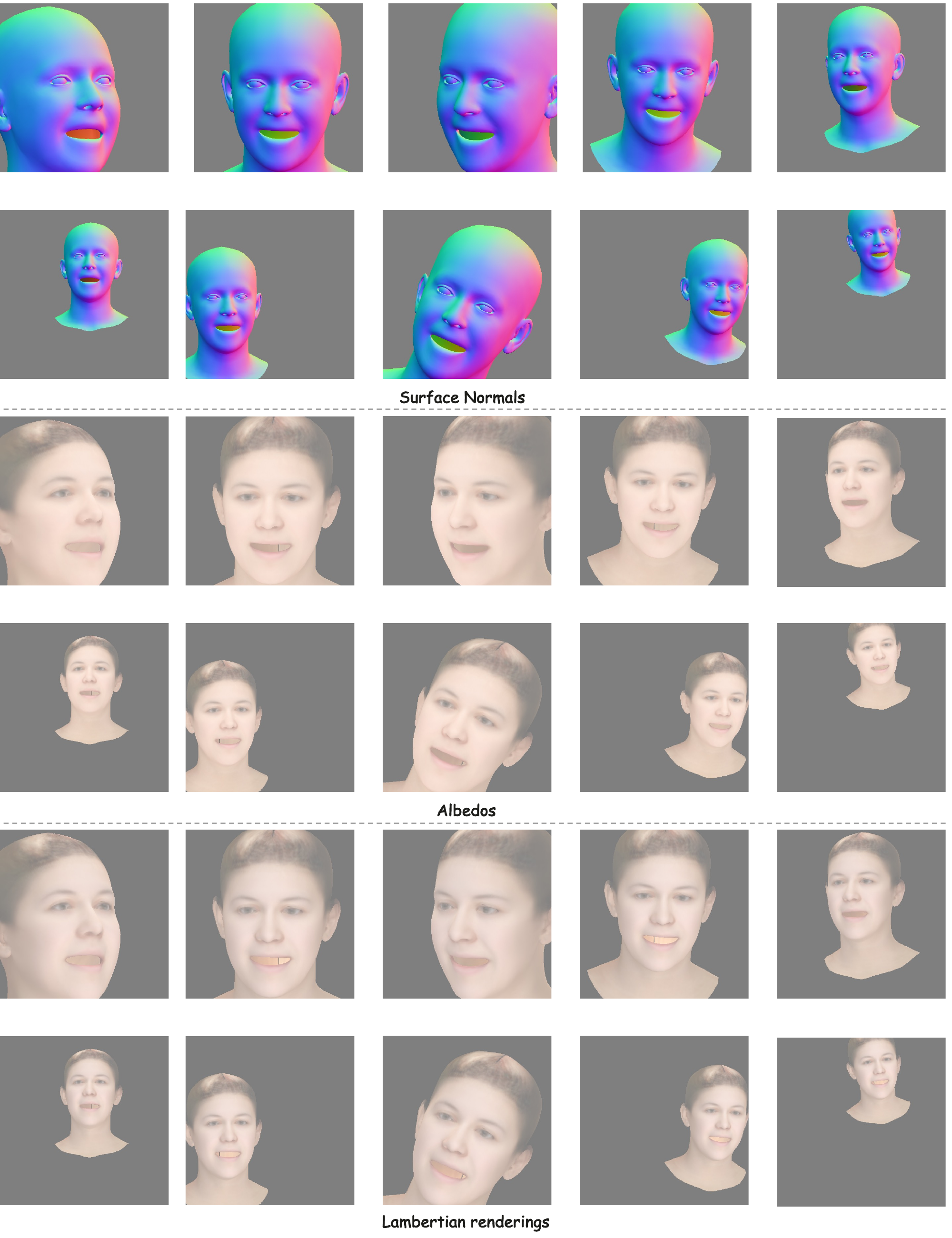}
\vspace{-1.9em}
\caption{
\textbf{Head Conditions.} We list the head conditions of a specific individual in our HumanIPHC benchmark, including Surface Normals, Albedos, and Lambertian renderings.
}
\label{fig:supp_benchmark}
\end{figure*}

\paragraph{Head conditions.} In Figure~\ref{fig:supp_benchmark}, we show the head conditions of a specific individual in our benchmark, including Surface Normals, Albedos, and Lambertian renderings.

\section{User Study Details}

We asked the participants to fill out the questionnaires. Every participant is required to score for each question. The score ranges from 1 to 5. The questions are listed as follows:
\begin{itemize}
    \item Given the reference image and generated image, score for the identity similarity. (1: pretty dissimilar, 5: pretty similar).
    \item Given the text prompt and generated image, score for the text-to-image alignment. (1: the image is pretty inconsistent with the text prompt, 5: the image is pretty consistent with the text prompt).
    \item Given the reference image, head condition, and generated image, score for the head control precision from the view of the shape, pose, position, lighting, and facial expression. (1: pretty bad, 5: pretty good).
    \item Given the generated image, score for the image quality. (1: pretty far away from the real image, 5: pretty close to the real image).
\end{itemize}

\section{Limitations and Social Impact}
\paragraph{Limitations.} Although our proposed method can achieve promising generative results, it still has several limitations. Our basic generative capabilities come from the pre-trained model, suggesting that our model might fail to generate the scenario out of the pre-training distribution. On the other hand, our 3D facial representation reconstruction relies on the estimation accuracy of DECA~\cite{DECA}. We find it struggles for some extreme poses and facial expressions. This can cause the misalignment of our generated images and the expected head conditions in some cases. Besides, the text richness is limited in our training data. It might be the reason that the text-to-image alignment performance degrades after training. Utilizing permissioned internet data might help alleviate this issue. We leave it for future research.

\paragraph{Social Impact.} Generative AI has drawn exceptional attention in recent years. Our research aims to provide an effective tool for human-centric image synthesis, especially for portrait personalization with head control in a flexible, fine-grained, and 3D-consistent manner. We believe it will play an important role in many potential entertainment applications. Like other existing generative methods, our method is susceptible to the bias from the large pre-trained dataset as well. Some malicious parties might have the potential to exploit this vulnerability for bad purposes. We encourage future research to address this concern. Besides, our model is at risk of abuse, \eg synthesizing politically relevant images. This risk can be mitigated by some deepfake detection methods~\cite{aghasanli2023interpretable, Corvi_2023_ICASSP} or by controlling the release of the model strictly.

%%%%%%%%%%%%%%%%%%%%%%%%%%
% if size > 20 MB, remove this figure
%%%%%%%%%%%%%%%%%%%%%%%%%%
% \begin{figure*}[t!]
% \centering
% \includegraphics[width=1.0\textwidth]{figures/fig-morecomp2-cam.pdf}
% \vspace{-1.9em}
% \caption{
% \textbf{More qualitative results.} Our CapHuman can produce well-identity-preserved, photo-realistic, and high-fidelity portraits with various head positions and poses in different contexts, compared with the baselines. 
% Note that our model can be combined with other pre-trained models, \eg RealisticVision~\cite{Realistic} in the community flexibly. For the head condition, we only display the Surface Normal here.
% }
% \label{fig:supp_main_results2}
% \end{figure*}

\begin{table*}\centering
\resizebox{.95\textwidth}{!}{
\begin{tabular}{c c c c c c c c c c }
$182723$&$182765$&$182828$&$182879$&$183243$&$183262$&$183344$&$183401$&$184642$&$184712$\\
$184713$&$184848$&$184858$&$184998$&$185120$&$185758$&$185827$&$186101$&$186436$&$186479$\\
$186538$&$186862$&$186981$&$187031$&$187083$&$187958$&$187990$&$188016$&$188082$&$188346$\\
$188646$&$189420$&$189454$&$189597$&$189635$&$189888$&$189913$&$189930$&$190093$&$190146$\\
$190971$&$190986$&$191153$&$191611$&$191663$&$191847$&$192006$&$192254$&$192279$&$192541$\\
$192816$&$192904$&$193230$&$193793$&$194155$&$194303$&$194309$&$194330$&$194629$&$194656$\\
$195350$&$195514$&$196047$&$196099$&$196205$&$196251$&$196475$&$196824$&$197119$&$197129$\\
$197168$&$197210$&$197464$&$197630$&$197829$&$198143$&$198223$&$198234$&$198413$&$198614$\\
$198869$&$198909$&$199377$&$199538$&$199621$&$199732$&$200305$&$200504$&$200505$&$201191$\\
$201546$&$201703$&$201731$&$201737$&$201915$&$201962$&$202244$&$202338$&$202459$&$202515$
\end{tabular} }
% \vspace{-0.7em}
\caption{\textbf{ID list.} We list all the IDs used in our HumanIPHC benchmark.
\label{tab:id_list}}
\end{table*}

\end{document}